Full Length Article

# Toward developing socially compliant automated vehicles: Advances, expert insights, and a conceptual framework

Yongqi Dong [a,b,*], Bart van Arem [a], Haneen Farah [a]

[a] Faculty of Civil Engineering and Geosciences, Delft University of Technology, Delft, 2628 CN, the Netherlands
[b] Institute of Highway Engineering, RWTH Aachen University, Aachen, 52074, Germany

## ARTICLE INFO

## ABSTRACT

By improving road safety, traffic efficiency, and overall mobility, automated vehicles (AVs) hold promise for revolutionizing transportation. Despite the steady advancement in high-level AVs in recent years, the transition to full automation entails a period of mixed traffic, where AVs of varying automation levels coexist with human-driven vehicles (HDVs). Making AVs socially compliant and understood by human drivers is expected to improve the safety and efficiency of mixed traffic. Thus, ensuring AVs' compatibility with HDVs and social acceptance is crucial for their successful and seamless integration into mixed traffic. However, research in this critical area of developing socially compliant AVs (SCAVs) remains sparse. This study carries out the first comprehensive scoping review to assess the current state of the art in developing SCAVs, identifying key concepts, methodological approaches, and research gaps. An informal expert interview was also conducted to discuss the literature review results and identify critical research gaps and expectations toward SCAVs. On the basis of the scoping review and expert interview input, a conceptual framework is proposed for the development of SCAVs. The conceptual framework is evaluated via an online survey targeting researchers, technicians, policymakers, and other relevant professionals worldwide. The survey results provide valuable insights and affirm the importance of the proposed conceptual framework in tackling the challenges of integrating AVs into mixed-traffic environments. Additionally, future research perspectives and suggestions are discussed, contributing to the research and development agenda of SCAVs.

## Nomenclature

| Abbreviation | Definition |
| --- | --- |
| AD | Automated driving |
| AI | Artificial intelligence |
| AV | Automated vehicle |
| CNN | Convolutional neural network |
| CSV | Comma-separated value |
| DL | Deep learning |
| DNN | Deep neural networks |
| eHMI | External human–machine interfaces |
| GAN | Generative adversarial network |
| GRU | Gated recurrent unit |
| HDV | Human-driven vehicle |
| IEEE | Institute of electrical and electronic engineers |
| IRL | Inverse reinforcement learning |
| LSTM | Long short-term memory neural network |
| ML | Machine learning |
| MLP | Multilayer perceptron |
| MDP | Markov decision process |

*(continued on next column)*

*(continued)*

| Abbreviation | Definition |
| --- | --- |
| MPC | Model predictive control |
| NGSIM | Next Generation Simulation |
| ODD | Operational design domain |
| OEM | Original equipment manufacturer |
| POMDP | Partially observable Markov decision process |
| POSG | Partially observable stochastic game |
| PPO | Proximal policy optimization |
| RL | Reinforcement learning |
| ROS | Robot operation system |
| SAE | Society of automotive engineers |
| SCAV | Socially compliant automated vehicle |
| SVO | Social value orientation |
| TRID | Transport research international documentation |
| V2V | Vehicle-to-vehicle |
| V2I | Vehicle-to-infrastructure |
| V2X | Vehicle-to-everything |

* Corresponding author. Faculty of Civil Engineering and Geosciences, Delft University of Technology, Delft, 2628 CN, the Netherlands.
  *E-mail address:* yongqi.dong@rwth-aachen.de (Y. Dong).

# 1. Introduction

Automated vehicles (AVs) are expected to benefit traffic safety and efficiency (Greenblatt and Shaheen, 2015; Jamson et al., 2011; Talebpour and Mahmassani, 2016; Yaqoob et al., 2020). Although the steady development of higher levels of AVs is gradually occurring, their deployment will not occur overnight. Instead, a transition period is inevitable, during which AVs with various automation levels will share the same road environment with human drivers, leading to mixed traffic conditions.

The society of automotive engineers (SAE) defines six levels of driving automation (SAE International, 2021), ranging from no driving automation (Level 0) to full driving automation (Level 5). Level 0 has no automation, and the driver is fully responsible for all aspects of driving. Levels 1 and 2 introduce partial automation, where the driver remains responsible for driving, even with the assistance of automated features, and must supervise these features continuously. The difference between Levels 1 and 2 lies in the scope of control supported: Level 1 supports either steering or braking/acceleration, whereas Level 2 supports both simultaneously, encompassing longitudinal and lateral control. At levels 3, 4, and 5, the automated system monitors the environment with full automation capabilities when the automated driving (AD) features are engaged. However, distinctions exist among these levels. At Level 3, known as conditional automation, drivers must be prepared to intervene and resume control when prompted by the AD features. However, at Levels 4 and 5, the AD features never make such requests. For Level 4, the AD features can operate the vehicle only under specific conditions defined by the operational design domain (ODD). In contrast, Level 5 allows the AD features to operate the vehicle under all conditions.

The deployment of AVs with varying levels of automation in mixed traffic introduces new challenges and novel interactions, which may create uncertainties and issues that affect both road safety and efficiency (Fagnant and Kockelman, 2015; Farah et al., 2022; Fraedrich et al., 2015; Raju et al., 2022). The uncertainties and challenges stem from the interaction between AVs and other vehicles at different levels of automation and HDVs in mixed traffic. Unlike fully automated environments, these interactions in mixed traffic introduce variability in driving behaviors, decision-making processes, and responses to dynamic road conditions. Factors such as various reaction times, unpredictable human behavior, and potential inconsistencies in AV decision models contribute to these uncertainties. Moreover, there is a pressing need to ensure the acceptance of AVs by human drivers to seamlessly integrate them into existing traffic systems (Łach and Svyetlichnyy, 2024; Orieno et al., 2024).

With respect to the development of AVs' driving behaviors, previous studies have traditionally prioritized aspects such as safety, efficiency, comfort, and energy consumption (Du et al., 2022; ElSamadisy et al., 2024; Vasile et al., 2023; Zhu et al., 2020). While these elements are essential, the growing complexity of mixed-traffic environments—where AVs must coexist with human-driven vehicles (HDVs)—highlights the importance of ensuring that AVs' driving behaviors are socially compliant. Referring to the definition provided in (Schwarting et al., 2019), socially compliant driving of AVs can be defined as behaving predictably and complying with the social expectations of human drivers and other surrounding road users (including other AVs) when they encounter social dilemmas while driving with intensive interactions (e.g., driving through unsignalized intersections, roundabouts, on-ramp/off-ramp merging, or unprotected left turning). This encompasses compliance with different local driving cultures, norms, cues, formal and informal traffic rules, and behaviors expected in specific contexts. The ability of AVs to drive in a predictable and socially compliant way is critical not only for enhancing safety and efficiency but also for fostering the understanding and acceptance of AVs by human drivers. Consequently, interest in designing and developing socially compliant automated driving systems is increasing. AVs with socially compliant driving capabilities, i.e., socially compliant AVs (SCAVs), generally correspond to Level 3 to Level 5 automation. While infrequent, certain aspects of socially compliant driving might also be observed at Level 2 or Level 1 automation, where partial driver assistance needs to be provided when requested. Nevertheless, the full potential of socially compliant AVs is most relevant and impactful at higher levels of automation, where AVs are expected to make independent decisions in complex traffic scenarios.

Some preliminary efforts have been made in the domain of socially compliant driving (Hang et al., 2021; Kolekar et al., 2020; Schwarting et al., 2019; Wang et al., 2022). These studies have laid important groundwork by exploring various aspects of the social compliance of AVs, including modeling social interactions, understanding the dynamics between HDVs and AVs, and developing models for socially aware perceptions, decision-making, or trajectory planning. However, despite these advancements, research on this emerging topic remains relatively limited, particularly in areas such as the modeling of different driving norms and implicit communication in different cultural backgrounds. Current studies lack a comprehensive, integrated approach that fully addresses the complex, multidisciplinary, and multifaceted nature of socially compliant driving. Therefore, there is a clear and pressing need for the development of an integrated conceptual framework that can guide future research, providing a holistic understanding of socially compliant driving and helping to design a research agenda to bridge the gaps in the current literature.

To advance research in the domain of SCAVs, this study embarks on a comprehensive approach employing an integrated research method. It begins with a scoping review of the current state of the art, aimed at identifying key concepts, methodological approaches, and research gaps. Additionally, an informal expert interview was conducted to gather insights into critical issues and research expectations for SCAVs. Subsequently, leveraging the findings from the scoping review and expert interviews, a conceptual framework is proposed. This framework incorporates all aspects deemed necessary, on the basis of the scoping review and expert interviews, for the development of SCAVs. To validate and refine the proposed conceptual framework and gain further insights, an online survey was developed, and responses from experts worldwide were collected. The survey results provide valuable validation and insights, affirming the significance of the framework for developing SCAVs to safely and efficiently integrate them into mixed-traffic environments. Additionally, suggestions for future enhancements are elicited, contributing to the continuous development of AV technology and guiding potential directions for further research and development.

# 2. Scoping literature review

In this study, a scoping review is adopted to synthesize the current research evidence and state of practice in scientific peer-reviewed publications, as well as to identify the key concepts, predominant research approaches, and research gaps related to SCAVs.

A scoping review was selected over a systematic review because of the exploratory nature of the research objective. Compared with systematic reviews, which aim to provide a synthesis and critical appraisal of the published evidence (Munn et al., 2018), scoping reviews are more suitable for summarizing and reporting research evidence on emerging and burgeoning topics, where evidence is limited and not yet systematically consolidated. As outlined in (Arksey and O'Malley, 2005; Tafidis et al., 2022), scoping reviews aim to provide a broad overview of available research, identifying relevant key concepts, methodologies, and gaps that require further investigation. Considering that AVs, especially SCAVs, are still in the early stages of development, with a relatively small body of research, a scoping review approach is more appropriate for mapping the current state of the field.

The scope of the review specifically targets methodologies and technical developments (i.e., the methods, algorithms, platforms, tools, and datasets that have been employed), as well as the substantive content of reviewed studies (e.g., what has been done, what scenarios/

maneuvers have been covered), that are relevant to SCAVs. This focus aligns with the study's goal of proposing a conceptual framework to guide future research and development. The descriptive nature of the scoping review allows for an expansive exploration of the research landscape, offering a foundation for conceptualizing SCAVs in the context of mixed-traffic environments. Importantly, detailed analyses and discussions of the findings and conclusions from the reviewed studies are beyond the scope of this study, as the primary focus is on synthesizing key methodological insights to inform the proposed framework.

### 2.1. Five-step approach

In this study, a five-step scoping review was utilized to identify and report related literature and map the results. The five steps of the methodological approach are as follows:

Step 1: Setting up eligibility criteria and information sources
Step 2: Developing the search strategy and process
Step 3: Screening and selecting studies
Step 4: Charting and visualizing the studies
Step 5: Summarizing, synthesizing, and reporting the results

This five-step approach is a condensed version of the well-designed Preferred Reporting Items for Systematic reviews and Meta-Analyses (PRISMA) Extension for Scoping Reviews (PRISMA-ScR) (Tricco et al., 2018), developed in consultation with an international panel of experts to enhance research and scientific publications.

#### 2.1.1. Step 1: Setting up eligibility criteria and information sources

In this step, eligibility criteria and information sources are established to guide the selection of studies for the scoping review. In principle, only peer-reviewed research papers published in journals and conference proceedings in English up until May 21, 2024, were considered eligible for the scoping review. It is essential that pertinent studies involve social interactions between AVs and HDVs or between AVs and other road users (e.g., cyclists and pedestrians). Publications solely discussing and modeling social interactions and behaviors among humans (e.g., drivers, cyclists, and pedestrians) without insights into SCAVs are deemed ineligible and thus excluded from the review process. There have been a few review papers, including such publications (Benrachou et al., 2022; Crosato et al., 2023b; Wang et al., 2022; Zhang et al., 2023b). Therefore, the main difference and key contribution of the literature review in this study lies in its dedicated focus on socially compliant driving, specifically emphasizing interactions involving AVs or insights toward this goal as a core criterion.

Various academic databases and repositories have been used, including Scopus, Web of Science All Databases (Web of Science for short) (not the Web of Science Core Collection), IEEE Xplore, and Transport Research International Documentation (TRID). These four databases provide access to a wide range of peer-reviewed research papers published in journals and conference proceedings, offering comprehensive coverage of scholarly literature in the field of transportation and automated driving research.

#### 2.1.2. Step 2: Developing the search strategy and process

In this step, a systematic search strategy is developed to identify relevant studies for inclusion in the scoping review. The search strategy encompasses a combination of keywords and controlled vocabulary terms related to socially compliant automated driving, social-aware automated driving, social interaction, automated driving, and other associated concepts. Recognizing the varied terminologies used in the domain of automated driving, the search includes different spellings, synonyms, and variants of related concepts to ensure inclusivity. The keywords for each associated term are illustrated in Table 1, facilitating a nuanced and exhaustive search process.

Boolean operators and truncation were used to increase the precision and comprehensiveness of the search. The employed search strings were tailored to meet the specific requirements (e.g., length) and functionalities of each selected database. The time range was set to 2000–2024. The language of the publications was limited to English. Furthermore, only the publications within the subject areas of mathematics, psychology, physics, neuroscience, computer science, behavioral sciences, social sciences, operations research and management science, engineering (transport, robotics, telecommunications, automation control systems, etc.), and science technology were considered valid. Publications falling into other domains, e.g., art, architecture, demography, international relations, public administration, and social issues, were excluded.

Importantly, the literature search was carried out in two phases. One phase occurred before the conceptual design and online questionnaire survey, and the second phase occurred afterwards to capture new publications that had emerged during that time period.

#### 2.1.3. Step 3: Screening and selecting studies

In this phase, the screening process commences with an initial evaluation of the titles, abstracts, and keywords of the search results to determine their alignment with the research objectives and relevance to the study topic. This preliminary assessment serves to identify potentially eligible studies for further consideration. The full-text articles of the identified studies subsequently underwent a thorough review to assess their eligibility. Only studies that are deemed truly pertinent to the research objectives are selected for inclusion in the scoping review.

Furthermore, to ensure the comprehensiveness of the literature coverage, a backward and forward snowballing technique was employed. This technique involves examining the reference lists of the selected papers and the papers that cite the selected papers to identify additional relevant studies that may have been missed in the initial search.

#### 2.1.4. Step 4: Charting and visualizing the studies

In this step, the selected studies undergo abstraction and charting to capture their general characteristics, including authorship details, year of publication, source of publication, disciplinary focus of the journal or conference, keywords, abstract content, number of citations, etc. This process enables a comprehensive overview of the literature landscape and facilitates the identification of trends, patterns, and relationships among the selected studies.

Furthermore, keyword network analysis via VOSviewer (van Eck and Waltman, 2010) and Sankey diagram visualization techniques were employed to visually represent the relationships among key terms of methodologies adopted and targeted use cases in the identified studies. Keyword network analysis provides insights into the interconnectedness of key terms and concepts within the literature, highlighting prominent themes and areas of focus. By analyzing the co-occurrence and relationships between keywords, researchers can identify clusters of related concepts and uncover overarching themes. Sankey diagram visualization offers a graphical representation of the flow of information between different categories or variables, illustrating the distribution and relationships between various elements in the selected studies and providing a holistic view of the research landscape. By visualizing the flow of information, researchers can identify patterns, trends, and relationships that may not be immediately apparent from textual analysis alone.

**Table 1**
Keywords used for each associated term.

| Term | Relevant keyword |
|---|---|
| Automated vehicle | (Autonomous OR automated OR driverless OR driver-less OR self-driving OR selfdriving) AND (car OR vehicle); (Autonomous OR automated) driving |
| Socially compliant driving | (Social OR social-aware OR socially compliant OR human-like) AND (driving OR interaction OR behavior OR behavior OR navigation OR decision-making OR trajectory planning OR planning and control); driving AND (social compliance OR social acceptance) |

By leveraging these visualization techniques, the findings of the scoping review are presented in a clear and concise manner, enabling stakeholders to easily interpret and understand the key findings and insights derived from the selected studies. Additionally, visualizing the data enhances the accessibility and communicability of the research findings and facilitates knowledge dissemination. Therefore, researchers can gain deeper insights into the structure and content of the literature, ultimately contributing to a more comprehensive understanding of the research field.

### 2.1.5. Step 5: Summarizing, synthesizing, and reporting the results

In this final step, the results of the scoping review are synthesized and mapped on the basis of the extracted and charted data, as well as the findings from keyword network analysis and Sankey diagram visualization. The synthesized results are organized into clusters highlighting key themes, methodological approaches, application cases, study designs, models, metrics used, and broad findings identified in the selected studies. This allows for the identification of commonalities and differences among studies and provides a comprehensive overview of the literature landscape.

Furthermore, relevant research gaps were identified on the basis of the synthesized results, highlighting areas where further investigation is needed and providing valuable insights for the development of an integrated conceptual framework that addresses key challenges and opportunities in the development of SCAVs.

### 2.2. Scoping literature review results

### 2.2.1. Selection of pertinent studies

The literature search through the four selected academic databases and under the aforementioned search process originally returned 1542 records, i.e., 432 records (361 published documents and 71 preprints) by Scopus, 258 records by Web of Science (publications and preprints together), 634 records by IEEE Xplore (including early access articles), and 218 records by TRID. Additionally, 11 studies that were identified during the screening process through snowballing were added so that, in total, 1553 studies were eligible for the screening process.

These records were exported as comma-separated value (CSV) files and processed via the pandas Python data analysis library to merge and group the records and remove duplicates. Together with the manual examination of the titles, a total of 1327 valid unique records were included in the preliminary checking process. Then, on the basis of the title and abstract, 209 studies were identified to be either directly relevant to or capable of providing valuable insights into automated driving interactions with HDVs in mixed traffic, among which four are review or survey papers (Benrachou et al., 2022; Crosato et al., 2023b; Wang et al., 2022; Zhang et al., 2023b), and one is about cognitive architecture design and perspectives (Xie et al., 2020). Following a detailed examination of their full texts, 68 articles were ultimately screened out because of their potential to contribute significantly to the understanding and development of socially compliant automated driving in mixed traffic. Thus, the 68 studies were ultimately selected for in-depth review. Fig. 1 illustrates the selection process of pertinent studies under the PRISMA pipeline. A full list of the 209 studies is provided in Supplementary Attachment 1 at https://lnkd.in/gpceU6gQ.

### 2.2.2. Charting, visualizing, summarizing, synthesizing, and reporting the results

First, to visualize the key terms, methods, and concepts related to socially compliant driving and the development of SCAVs, the relevant publications identified by the Web of Science search engine were visualized via the keyword network plot in VOSviewer shown in Fig. 2. This study selected Web of Science as the sole database for visualization because of VOSviewer's limitations and the practical challenges associated with integrating multiple databases. The Web of Science database effectively captures the primary information and relationships between key terms and concepts, making it a suitable choice for constructing keyword network visualization. The size of the nodes and thickness of the links depict the scale of the publications in the corresponding areas of the keyword, and the different colors depict the clusters.

The analysis shows that decision-making appears to be the most frequent keyword, followed by terms such as agent, policy, dataset, robotics, human driver, robots, safety, and efficiency, among others. From visualizations, one can also identify commonly adopted methods and terms, such as deep learning (DL), neural networks, game theory, model predictive control (MPC), and optimization. These results provide valuable global insights for understanding the target domain of socially compliant driving.

To design SCAVs, methodologies identified in the reviewed literature can be broadly grouped into learning-based and model/utility-based approaches. In practice, learning-based and model-based approaches often complement each other to achieve more robust and adaptable performance. Specifically, the detailed methodologies can be roughly further classified into five key subcategories:

1) Imitation learning of social driving behaviors from human drivers

This approach focuses on replicating the social driving behaviors of human drivers through imitation learning techniques, such as behavior cloning (Wang et al., 2023b, 2023c; Zhu and Zhao, 2023), inverse reinforcement learning (IRL) (Geng et al., 2023; Sun et al., 2019), and generative adversarial imitation learning, e.g., in Da and Wei (2023). The AV learns to mimic human-like decision-making and driving patterns by observing and imitating either expert demonstrations or processed empirical real-world driving data. This method can work in an end-to-end pipeline but is not necessary. Representative works in this direction include (Da and Wei, 2023; Sun et al., 2019; Wang et al., 2021b; Xu et al., 2023).

2) Reinforcement learning (RL) combined with utility-based models

In this approach, RL is employed to infer the underlying utility (also referred to as reward in many studies) functions that govern social driving behaviors from observed human (expert) demonstrations or empirical driving data. The utility functions quantify social factors such as deterministic courtesy (Sun et al., 2018) and the magnitude of the concern people have for others relative to themselves, e.g., through social value orientation (SVO) (Liebrand and McClintock, 1988; Murphy and Ackermann, 2014; Schwarting et al., 2019). This method enables AVs to learn and adapt the relevant social factors influencing human decision-making to achieve socially compliant behavior (Buckman et al., 2019; Larsson et al., 2021; Nan et al., 2024; Schwarting et al., 2019; Wang et al., 2021a; Xue et al., 2023; Yoon and Ayalew, 2019).

3) Model-based generation of human-like behaviors

This category encompasses approaches that leverage mathematical models to replicate human driving behaviors and/or inform socially aware decision-making. Techniques within this category, such as game theory, social force models, driving risk field models, and potential field models, simulate the complex interaction dynamics between AVs and other road users, including HDVs, pedestrians, and cyclists. Game theory, in particular, provides a framework for strategic decision-making by modeling interactions as a series of cooperative or competitive scenarios where AVs make decisions on the basis of anticipated responses from surrounding agents (Hang et al., 2021, 2022c; Shu et al., 2023). Other models, such as the social force model (Chen et al., 2024; Reddy et al., 2021; Yoon and Ayalew, 2019), the driving risk field model (Geng et al., 2023; Kolekar et al., 2020; Wang et al., 2023a), and the potential field model (Bhatt et al., 2022; Yan et al., 2022; Zhao et al., 2024), capture the forces, risks, and potential outcomes of interactions in mixed-traffic environments, allowing for a more nuanced emulation of human-like behaviors. These model-based approaches are valuable for predicting and generating socially compliant driving behaviors by considering both explicit rules and inferred human tendencies. Notable contributions in this area include Bhatt et al. (2022); Ferrer and Sanfeliu (2014); Hang et al. (2021); Kolekar et al. (2020); Liu et al. (2024a); Shu et al. (2023); Wang et al. (2023a); Zhang et al. (2023a), etc.

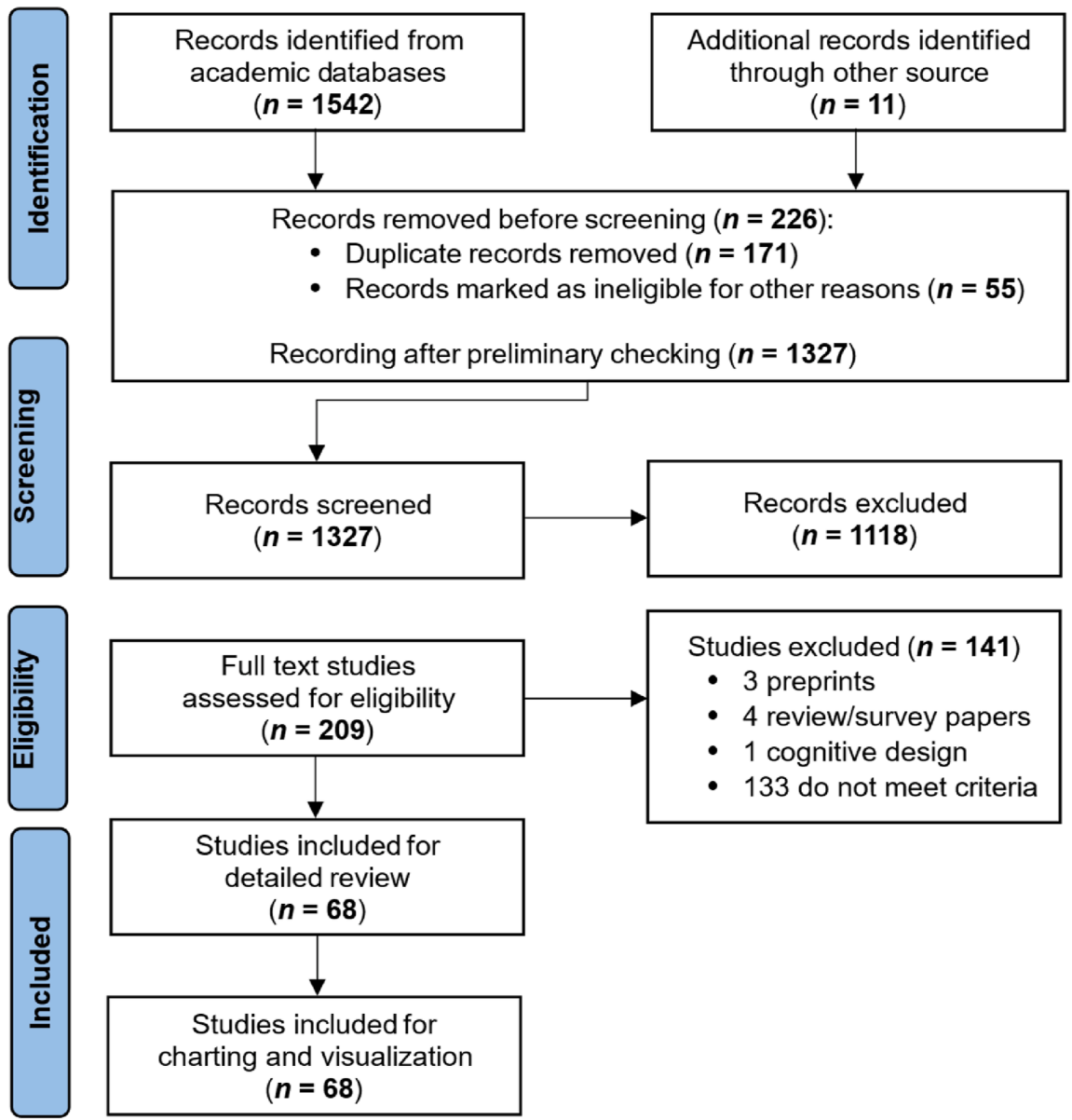

**Fig. 1.** Flow diagram of the process of selecting pertinent studies.

These models can usually be integrated with learning-based approaches (especially RL) to increase their adaptability and responsiveness in real-time applications, as seen in works such as (Liu et al., 2024a; Wang et al., 2024b).

4) Trajectory prediction through the integration of social factors with machine learning (ML) for the promotion of socially compliant behaviors

This subcategory focuses on the use of ML models integrated with social factors to predict trajectories that reflect socially compliant behavior. Unlike categories (1) and (2), which generally deliver driving control actions, the approaches here rely on DL via deep neural networks (DNNs) or IRL aided by social factor models to analyze and learn from large datasets and forecast the socially compliant trajectories of surrounding HDVs, pedestrians, and/or other road users. By accurately predicting these trajectories, the ego AV can then adjust its actions to achieve corresponding socially compliant driving behavior, thus ensuring smoother and safer interactions in mixed-traffic scenarios (Geng et al., 2023; Vemula et al., 2018; Yoon and Ayalew, 2019). The prediction can then be used for RL control (Valiente et al., 2024) to leverage prediction and social awareness in RL decision-making to improve safety and efficiency.

5) Optimization-based tuning of social driving parameters

This approach leverages optimization techniques to fine-tune the parameters of driving models to achieve desired social objectives, such as individualistic, altruistic, or prosocial driving behavior. By adjusting and optimizing these parameters, the models aim to balance the trade-offs between safety, efficiency, and comfort while considering the benefits of the ego AV versus surrounding vehicles or other road participants in mixed-traffic environments. Representative studies in this category include, e.g., Larsson et al. (2021).

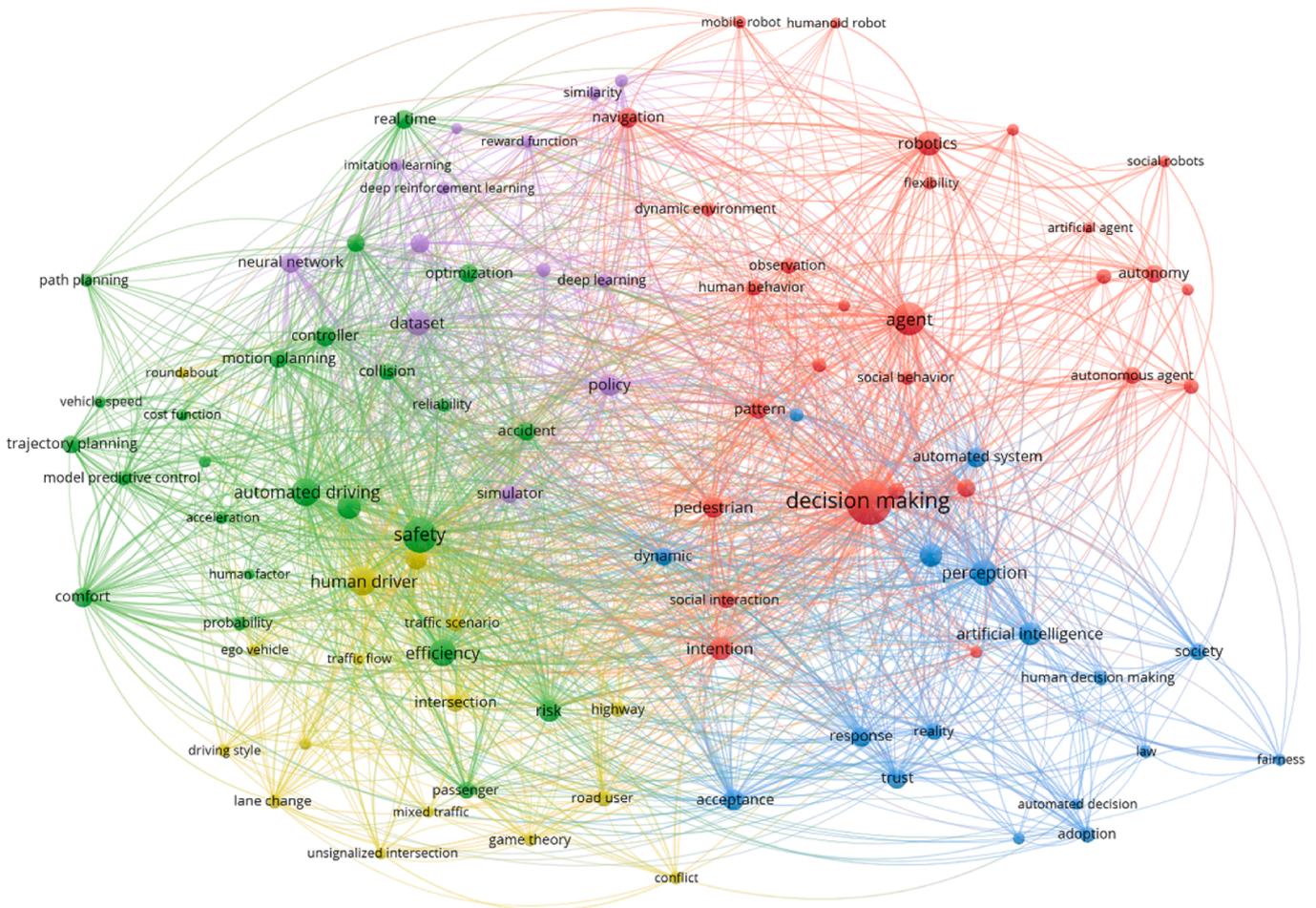

**Fig. 2.** Keyword network visualization by VOSviewer.

To provide a holistic view of the five identified methodological categories, Table 2 presents a comparative overview of their key characteristics, including advantages, disadvantages, and typical applications. This comparison elucidates the trade-offs and appropriate contexts for each approach, facilitating a deeper understanding of their roles in SCAV development.

As depicted in Table 2, each category offers distinct strengths and faces specific challenges. Imitation learning excels at replicating human behavior but is constrained by the diversity and quality of the available training data. Reinforcement learning offers adaptability to complex, dynamic settings, yet its effectiveness hinges on well-designed reward functions. Model-based approaches provide interpretable and theoretically sound frameworks for understanding interactions, such as through game theory, although they often demand significant computational resources and may struggle with adaptability and real-time application. Trajectory prediction, enriched by social factors, improves the anticipation of other road users' movements but depends on robust social data. Finally, optimization-based tuning allows precise adjustments to the driving parameters, although it may miss nuanced, dynamic social cues. Notably, these methodologies are complementary; combining them can harness their respective strengths to develop more robust and effective SCAV systems.

These aforementioned methodologies collectively represent the current state of research on socially compliant driving behavior for AVs. They highlight the multidisciplinary nature of the field, which combines

**Table 2**
Comparison of the five identified methodological categories.

| Methodological category | Advantage | Disadvantage | Typical application |
|---|---|---|---|
| Imitation learning of social driving behaviors from human drivers | Effectively replicates human behavior | Limited by diversity and quality of training data | Learning social driving norms from expert demonstrations |
| Reinforcement learning combined with utility-based models | Highly adaptable to dynamic environments | Sensitive to reward function design | Optimizing long-term socially compliant behavior |
| Model-based generation of human-like behaviors | Structured, interpretable, and theoretically grounded | Computationally intensive; may lack adaptability and real-time feasibility | Modeling strategic multiagent interactions (e.g., using game theory) |
| Trajectory prediction through integration of social factors with machine learning | Enhances prediction accuracy with social context | Reliant on the quality and availability of social data | Anticipating movements of HDVs or pedestrians |
| Optimization-based tuning of social driving parameters | Offers precise control over driving parameters | May overlook dynamic or implicit social cues | Fine-tuning AV responses for specific social scenarios |

elements of artificial intelligence (AI) (e.g., ML, DL, and RL), physics, human factors, control theory, social psychology, and transportation engineering. The integration of multidisciplinary knowledge is crucial for developing AVs capable of safely and efficiently interacting with HDVs and other road users in complex traffic environments. Importantly, the different approaches categorized are not mutually exclusive: In practice, they can be utilized in combination to increase the robustness and reliability of AV behavior. A detailed illustration of the models,

**Table 3**
Clustering of methods identified in the papers reviewed.

**(A) ML-based methods**

| Methods and terms adopted | | | Related publications |
|---|---|---|---|
| ML (DL, RL) | DL | CNN | Ding et al. (2022); Hirose et al. (2024); Pérez-Dattari et al. (2022); Qin et al. (2021); Valiente et al. (2024) |
| | | GAN | Da and Wei (2023); Gupta et al. (2018); Kothari and Alahi (2023); Sadeghian et al. (2019); Wang et al. (2021b) |
| | | LSTM | Alahi et al. (2016); Chang et al. (2023); Da and Wei (2023); Ding et al. (2022); Gupta et al. (2018); Huang et al. (2023a); Kothari et al. (2021); Kothari and Alahi (2023); Pérez-Dattari et al. (2022); Sadeghian et al. (2019); Vemula et al. (2018); Wang et al. (2024c); Wang et al. (2021b) |
| | | MLP | Chang et al. (2023); Da and Wei (2023); Huang et al. (2023a); Kothari and Alahi (2023); Xue et al. (2023); Zhu and Zhao (2023) |
| | | Transformer | Geng et al. (2023); Huang and Sun (2023); Huang et al. (2023a); Wang et al. (2024b) |
| | | Attention module | Kothari and Alahi (2023); Liu et al. (2024b); Qin et al. (2021); Sadeghian et al. (2019); Vemula et al. (2018); Wang et al. (2021b); Xue et al. (2023) |
| | | Graph attention network | Wang et al. (2024c) |
| | | Autoencoder | Liu et al. (2024b); Valiente et al. (2024); Zong et al. (2023) |
| | | GRU | Liu et al. (2024b); Zong et al. (2023) |
| | | Social pooling layer | Alahi et al. (2016); Gupta et al. (2018) |
| | RL | Actor-critic | Crosato et al. (2021); Crosato et al. (2023a); Huang et al. (2023b); Liu et al. (2024b); Liu et al. (2020); Toghi et al. (2021a); Tong et al. (2024); Xue et al. (2023); Zong et al. (2023) |
| | | Deep | Huang et al. (2023b); Lu et al. (2022); Nan et al. (2024); Taghavifar and Mohammadzadeh (2025); Toghi et al. (2021b, 2022); Valiente et al. (2024) |
| | | Q-learning | Valiente et al. (2024) |
| | | IRL | Geng et al. (2023); Huang et al. (2023a); Nan et al. (2024); Schwarting et al. (2019); Sun et al. (2018, 2019); Xu et al. (2023); Zhao et al. (2024) |
| | | Proximal policy optimization (PPO) | Crosato et al. (2023a); Liu et al. (2024b) |
| | | Coordinated policy optimization | Peng et al. (2021) |

**(B) Game theory, field-based models, and social psychological factor-related methods**

| Methods and terms adopted | | Related publication |
|---|---|---|
| Game theory | Stackelberg game | Hang et al. (2020, 2021, 2022b); Li et al. (2022); Schwarting et al. (2019); Wang et al. (2021a); Zhao et al. (2024) |
| | Nash-equilibrium based game | Galati et al. (2022); Hang et al. (2021), 2022b); Liu et al. (2024a, 2024c); Shu et al. (2023); Wang et al. (2023a) |
| | POSG | Toghi et al. (2021b, 2022); Valiente et al. (2024); Xue et al. (2023) |
| | Coalitional game | Hang et al. (2022a, 2022c) |
| | Potential game | Liu et al. (2024c) |
| Social psychological factor | SVO | Buckman et al. (2019); Crosato et al. (2021); Crosato et al. (2023a); Peng et al. (2021); Schwarting et al. (2019); Taghavifar and Mohammadzadeh (2025); Toghi et al. (2021a, 2021b, 2022); Tong et al. (2024); Valiente et al. (2024); Xue et al. (2023); Zhang et al. (2023a); Zhao et al. (2024) |
| | Courtesy | Chang et al. (2023); Li et al. (2022); Sun et al. (2018); Wang et al. (2021a) |
| | Coordination tendency | Liu et al. (2024b) |
| | Social preference | Lu et al. (2022) |
| | Social cohesion | Landolfi and Dragan (2018) |
| | Social anchor | Kothari et al. (2021) |
| Field-based models | Potential field | Bhatt et al. (2022); Hang et al. (2020, 2021, 2022a, 2022b); Reddy et al. (2021); Yan et al. (2022); Zhao et al. (2024) |
| | Risk field | Geng et al. (2023); Kolekar et al. (2020); Wang et al. (2023a); Wang et al. (2024b); Zhang et al. (2023a) |

**(C) Other models and methods**

| Methods and terms adopted | Related publication |
|---|---|
| Model predictive control | Bhatt et al. (2022); Hang et al. (2020, 2021, 2022a, 2022c); Landolfi and Dragan (2018); Larsson et al. (2021); Pérez-Dattari et al. (2022); Sun et al. (2018), 2019); Wang et al. (2021a, 2023a); Yan et al. (2022); Yoon and Ayalew (2019); Zhang et al. (2023a) |
| Markov decision process | Crosato et al. (2021, 2023a); Da and Wei (2023); Ding et al. (2022); Huang et al. (2023b); Liu et al. (2024b); Peng et al. (2021); Song et al. (2016); Zong et al. (2023) |
| Expert demonstration | Da and Wei (2023); Huang et al. (2023a, 2023b); Liu et al. (2024a); Nan et al. (2024); Qin et al. (2021); Zhu and Zhao (2023) |
| Social force model | Chen et al. (2024); Crosato et al. (2023a); Ferrer and Sanfeliu (2014); Reddy et al. (2021); Yoon and Ayalew (2019) |
| Addressing uncertainties | Huang et al. (2023b); Kolekar et al. (2020); Sun et al. (2019); Wang et al. (2021a) |
| Bayesian inference | Li et al. (2022); Wang et al. (2021a, 2023a) |
| Behavior cloning | Wang et al. (2023b, 2023c) |
| Monte-carlo sampling | Wang et al. (2023b, 2023b) |
| Monte carlo tree search | Li et al. (2022) |
| Finite state machine | Wang et al. (2024a) |
| Reasoning graph | Zhou et al. (2022) |
| Non-convex mixed-integer nonlinear program | Larsson et al. (2021) |
| Discrete choice model | Kothari et al. (2021) |
| Minimizing counterfactual perturbation | Hirose et al. (2024) |
| Particle filtering | Xu et al. (2023) |
| Gaussian process | Valiente et al. (2024) |
| Genetic algorithm | Liu et al. (2024a) |

terms, and methods adopted by the studies is provided in Table 3 and Fig. 3.

Table 3 shows that the majority of studies adopt ML approaches, more specifically, DL (e.g., convolutional neural network (CNN), generative adversarial network (GAN), long short-term memory (LSTM) neural network, multilayer perceptron (MLP), gated recurrent unit (GRU), transformer), and deep reinforcement learning (e.g., IRL, deep Q-learning, and actor-critic methods). Typically, driving decision-making is modeled as the Markov decision process (MDP) (Crosato et al., 2021; Da and Wei, 2023; Ding et al., 2022; Hang et al., 2021; Huang et al., 2023a; Liu et al., 2024b; Zong et al., 2023), or as the partially observable Markov decision process (POMDP) (Ding et al., 2022; Liu et al., 2024b; Peng et al., 2021), to account for uncertainties. Additionally, many studies employ game theory (e.g., the Stackelberg game, coalitional game, potential game, and partially observable stochastic game (POSG)) to effectively model complex interactions between agents (e.g., AVs and HDVs), whereas a significant portion also utilizes MPC to refine and smooth control outputs following decision-making. In the realm of social preferences, a variety of social psychological terms—such as courtesy, coordination tendency, and SVO—are used to encapsulate concepts related to social preferences. The targeted research objectives and tasks typically fall into three primary categories: behavior generation, trajectory prediction, and interactive decision-making and control. Furthermore, multiple-agent modeling has been incorporated in some studies to simulate complex, interactive driving environments involving multiple road participants (Da and Wei, 2023; Peng et al., 2021; Toghi et al., 2021a, 2021b, 2022; Xue et al., 2023). These observations align with insights from the keyword network visualization in Fig. 2 and are illustrated further in Fig. 3.

Furthermore, while some interdisciplinary initiatives have been introduced, the majority of research continues to focus on combining approaches from computer science, physics, mathematics, transportation, and vehicular engineering. Although initial efforts to incorporate social psychology are emerging, they focus primarily on concepts such as SVO, coordination tendencies, and courtesy, which share common themes. The development of more advanced models grounded in social psychology and other relevant interdisciplinary fields is essential to deepen the understanding of human-AV interactions (Brown et al., 2023; Vinkhuyzen and Cefkin, 2016). Specifically, incorporating culturally sensitive social behaviors into AV decision-making to develop customized AVs for diverse cultural backgrounds remains a crucial area for further investigation (Dong et al., 2024).

Table 4 groups the reviewed papers on the basis of simulation, data-driven, and empirical field testing approaches. Table 4 and Fig. 3 show that more than half of the studies employed simulations to train, test, and verify their solutions. The commonly adopted simulation platforms and software tools include Highway-env (Leurent, 2018), SMARTS (Zhou et al., 2020), CARLA (Dosovitskiy et al., 2017), MetaDrive (Li et al., 2023), PTV VISSIM, SUMO(Lopez et al., 2018), Universe simulator (Zhang, 2023), and Robot Operation System (ROS). Additionally, more than half of the studies incorporated empirical datasets collected from real-world environments to enhance model validation. Typical frequently used datasets include the next-generation simulation (NGSIM) dataset (U.S. Department of Transportation Federal Highway Administration, 2016), Waymo Open Motion dataset (Ettinger et al., 2021), INTERACTION dataset (Zhan et al., 2019), highD dataset (Krajewski et al., 2018), exiD dataset (Moers et al., 2022), inD dataset (Bock et al., 2020), rounD dataset (Krajewski et al., 2020), SinD dataset (Xu et al., 2022), Argoverse Motion dataset (Chang et al., 2019), and Argoverse 2 Motion dataset (Wilson et al., 2023). Additional datasets, including the ETH (Pellegrini et al., 2009), UCY (Lerner et al., 2007), TrajNet++ (Kothari et al., 2022), PANDA (Wang et al., 2020), Stanford Drone (Robicquet et al., 2016), and HuRoN (Hirose et al., 2024) datasets, are employed for scenarios and applications related to social robot navigation and human trajectory prediction.

Furthermore, as clearly illustrated in Table 5 and Fig. 4, regarding driving maneuvers, the majority of studies focus on those that require both longitudinal and lateral control. Various maneuvers, e.g., driving through unsignalized intersections, performing unprotected left turns, lane changing, on-ramp merging, and overtaking, have been studied. The inherent complexity and dynamic nature of these scenarios, where both directional and speed-related aspects of control must be simultaneously managed, make them particularly well suited for studying and examining social interactions between AVs and HDVs. Such scenarios provide robust "environments" for developing and validating socially compliant driving behaviors, as they compel AVs to navigate nuanced interactions, accommodate unpredictable human behaviors, convey their intentions, and adapt their decisions to align with various human social driving patterns. Interestingly, within the reviewed publications, only two studies specifically delve into maneuvers involving only longitudinal control, i.e., car-following. This may stem from the fact that longitudinal maneuvers are often already embedded within the broader, more complex scenarios mentioned above, and there is no need to specifically target only longitudinal maneuvers.

From Table 5 and Fig. 4, it is also important to note that some studies focus primarily on social robot navigation and human trajectory forecasting for related applications, with 12 studies included in the review. While AVs can be considered a type of robot and the insights from social robot navigation research could be beneficial for developing socially compliant driving, there are notable differences between human/pedestrian–robot interactions and the interactions between HDVs and AVs. These differences stem from the distinct speeds, operational environments, and interaction dynamics between the two scenarios. Social robot navigation often occurs at lower speeds and in more controlled environments, which facilitates the use of field test experiments to observe and refine socially aware behaviors. Insights gained from such experiments could serve as a foundation for adaptation to the more complex and high-speed interactions involved in AV driving. This study highlights some typical works related to pedestrian trajectory prediction and social robots navigating around humans but does not aim to provide a comprehensive review of these domains. For further information, readers are encouraged to refer to Singamaneni et al. (2024).

Finally, some papers delve into the public's social perception, acceptance, and trust of AV technology, e.g., Joo and Kim (2023); Oliveira et al. (2019); Othman (2021); and Schneble and Shaw (2021), recognizing these aspects as critical for the broader adoption and integration of AVs into society. In particular, Joo and Kim (2023) conducted an online study to explore the influence of perceived collision algorithm types, i.e., selfish (prioritizing passenger safety) versus utilitarian (minimizing total damage by saving more lives, regardless of passenger status), and the role of social approval of these algorithms on individuals' attitudes toward AVs. The study revealed a striking mismatch between societal and individual preferences. The participants rated utilitarian algorithms as more ethical and beneficial to society, aligning with broader social values. However, they expressed greater trust in, and a stronger personal preference for, selfish algorithms. The respondents were more willing to use and even pay a premium for AVs equipped with selfish algorithms, highlighting a significant divergence between ethical ideals and personal safety priorities. This discrepancy underscores the complexity of fostering public trust and acceptance of AV technology and suggests that designing and deploying SCAVs to balance societal ethics with individual user preferences is a crucial challenge for manufacturers and policymakers.

## 3. Conceptual framework design

### 3.1. Discussing the literature review results with experts

Building on the findings from the summarized literature review, an informal interview was conducted with ten experts representing diverse scientific and consultancy positions across research institutes,

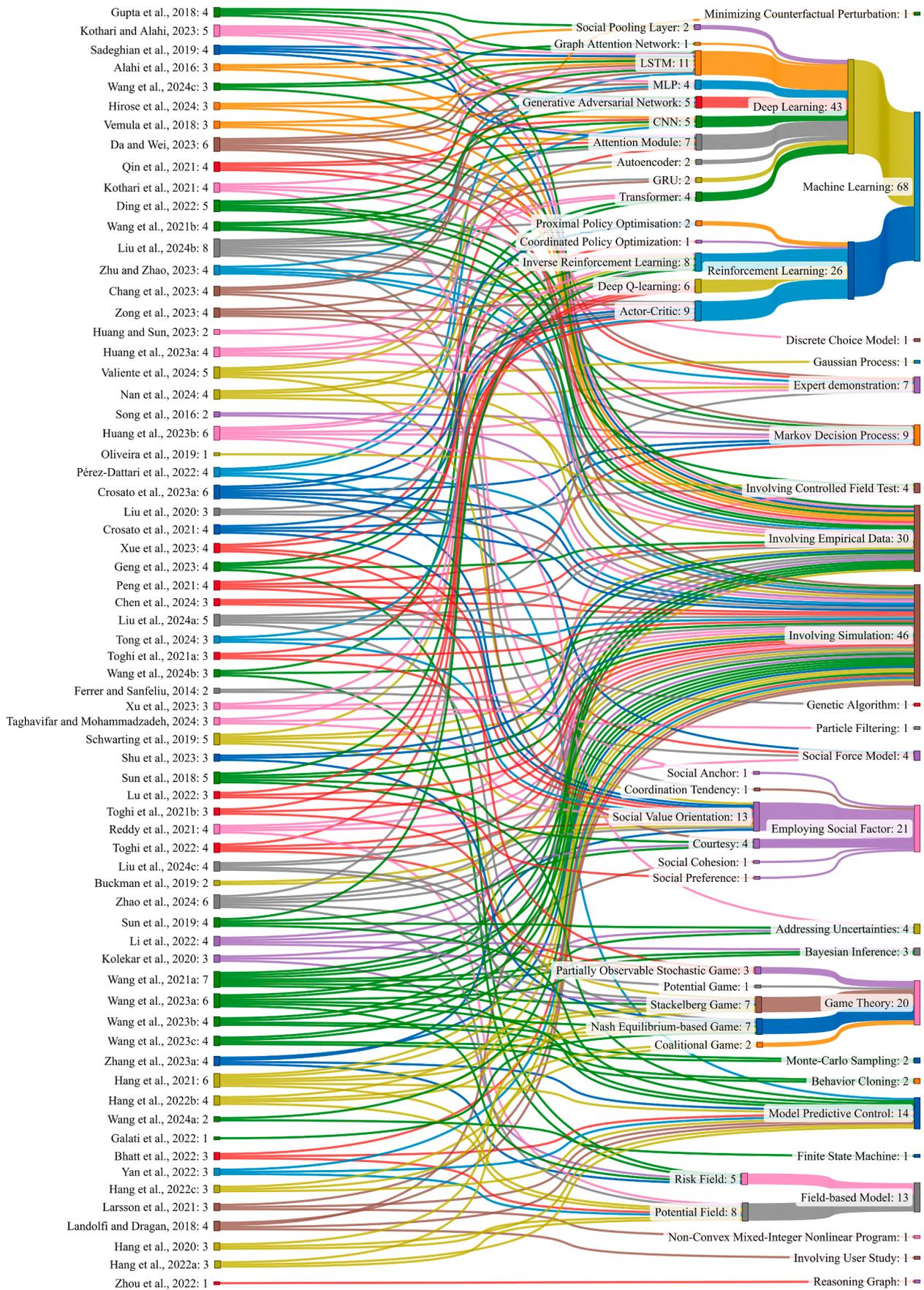

**Fig. 3.** Identified methods adopted in each study. Note: A single paper may involve multiple methods (e.g., both deep learning and reinforcement learning) and may utilize multiple models within the same method category (e.g., both LSTM and CNN within the DL category).

**Table 4**

Grouping of reviewed papers on the basis of simulation, data-driven, and empirical field testing approaches.

| Methods adopted | Tool, platform, or dataset | | Related publication |
|---|---|---|---|
| Simulation and simulator-related | Highway-env | | Liu et al. (2024b); Toghi et al. (2021a, 2021b, 2022); Tong et al. (2024); Valiente et al. (2024); Zhang et al. (2023a) |
| | SMARTS | | Huang et al. (2023b); Wang et al. (2024b) |
| | CARLA | | Bhatt et al. (2022); Lu et al. (2022); Pérez-Dattari et al. (2022); Zhu and Zhao (2023) |
| | MetaDrive | | Peng et al. (2021) |
| | Python-based | | Crosato et al. (2021, 2023a); Da and Wei (2023); Liu et al. (2020); Wang et al. (2021b) |
| | Python-MATLAB | | Zhao et al. (2024) |
| | MATLAB-Simulink | | Hang et al. (2020, 2021, 2022a, 2022c) |
| | Prescan | | Song et al. (2016) |
| | CarSim | | Chen et al. (2024) |
| | MATLAB/Simulink-CarSim | | Yan et al. (2022) |
| | Prescan-MATLAB/Simulink-CarSim | | Wang et al. (2023a) |
| | Robot operation system (ROS) | | Pérez-Dattari et al. (2022); Wang et al. (2021a) |
| | SUMO-ROS | | Zong et al. (2023) |
| | PTV VISSIM | | Larsson et al. (2021) |
| | Julia | | Sun et al. (2018) |
| | MobileSim | | Reddy et al. (2021) |
| | Universe simulator | | Xue et al. (2023) |
| | Fixed based driving simulator | | Kolekar et al. (2020) |
| | Human-in-the-loop driver simulator | | Liu et al. (2024a); Xu et al. (2023) |
| | Hardware-in-the-loop simulator | | Hang et al. (2022b) |
| | Self-built upon datasets | | Wang et al. (2023b) |
| | Not specified | | Buckman et al. (2019); Ferrer and Sanfeliu (2014); Landolfi and Dragan (2018); Schwarting et al. (2019); Shu et al. (2023); Sun et al. (2019); Taghavifar and Mohammadzadeh (2025); Wang et al. (2024); Zhou et al. (2022) |

| Methods adopted | Tool, platform, or dataset | | Related publication |
|---|---|---|---|
| Involving empirical data | Next generation simulation (NGSIM) dataset | | Chen et al. (2024); Hang et al. (2021); Liu et al. (2024c); Nan et al. (2024); Schwarting et al. (2019); Sun et al. (2018); Wang et al. (2023a); Zhao et al. (2024) |
| | Waymo open motion dataset | | Chang et al. (2023); Huang et al. (2023a) |
| | INTERACTION dataset | | Huang and Sun (2023); Li et al. (2022); Shu et al. (2023); Tong et al. (2024); Wang et al. (2021a); Wang et al. (2023c) |
| | highD dataset | | Wang et al. (2023b); Xu et al. (2023) |
| | exiD dataset | | Wang et al. (2023b) |
| | inD dataset | | Geng et al. (2023); Wang et al. (2023c) |
| | rounD dataset | | Wang et al. (2023c) |
| | SinD dataset | | Liu et al. (2024a) |
| | Argoverse motion dataset | | Ding et al. (2022) |
| | Argoverse2 motion dataset | | Liu et al. (2024a) |
| | Beijing Jianguomen flyover area dataset | | Wang et al. (2021b) |
| | Data collected by wheelchair testbed | | Qin et al. (2021) |
| | Data collected over 60 h of driving from 10 drivers at 6 intersections | | Zhu and Zhao (2023) |
| | Datasets related to social robot navigation/human trajectory prediction | PANDA | Wang et al. (2024c) |
| | | ETH | Alahi et al. (2016); Gupta et al. (2018); Kothari and Alahi (2023); Sadeghian et al. (2019); Vemula et al. (2018); Wang et al. (2024c) |
| | | UCY | Alahi et al. (2016); Gupta et al. (2018); Kothari and Alahi (2023); Sadeghian et al. (2019); Vemula et al. (2018); Wang et al. (2024c) |
| | | TrajNet++ | Kothari et al. (2021); Kothari and Alahi (2023) |
| | | Stanford Drone Dataset | Sadeghian et al. (2019) |
| | | HuRoN | Hirose et al. (2024) |
| | Involving controlled field test | | Ding et al. (2022); Ferrer and Sanfeliu (2014); Hirose et al. (2024); Liu et al. (2020); Oliveira et al. (2019); Reddy et al. (2021) |
| | Involving survey questionnaire | | Galati et al. (2022) |
| | Involving user study | | Landolfi and Dragan (2018) |

consulting firms, original equipment manufacturers (OEMs), and government sectors. The purpose of the interviews was to gather expert perspectives through open-ended discussions on the current limitations of AVs, to further identify existing research gaps and to understand their expectations for the development of SCAVs.

To facilitate insightful and meaningful discussions, the preliminary findings from the literature review were shared with the experts prior to the discussion. This ensured that the conversations were well informed. The discussions were open-ended, allowing participants to elaborate on their views on the current limitations of AVs and provide in-depth observations of the challenges and opportunities in this field. The questions

discussed include the following:

- Do you have confidence in automated vehicles, particularly in mixed-traffic conditions?
- What are the current limitations and critical pain points of automated vehicles?
- Which scenarios do you perceive as particularly challenging for automated vehicles, and what scenarios, maneuvers, or use cases would you like automated vehicles to address soon?
- What are your expectations for the short-term and long-term development of automated vehicles?

**Table 5**
Clustering of maneuvers and applications identified in the reviewed papers.

| Use case | | Related publication |
|---|---|---|
| Intersection | Unsignalized intersection[a] | Buckman et al. (2019); Geng et al. (2023); Hang et al. (2022a); Liu et al. (2024a, 2024c); Peng et al. (2021); Song et al. (2016); Valiente et al. (2024); Xia et al. (2022); Zhu and Zhao (2023); Zong et al. (2023) |
| | Unprotected left turn[a] | Hang et al. (2022b); Huang et al. (2023b); Liu et al. (2024a, 2024b); Schwarting et al. (2019); Shu et al. (2023); Wang et al. (2024b); Zhou et al. (2022); Zong et al. (2023) |
| | Roundabout | Huang et al. (2023b); Li et al. (2022); Peng et al. (2021); Valiente et al. (2024); Wang et al. (2021a); Zhang et al. (2023a) |
| | T-junction | Oliveira et al. (2019); Pérez-Dattari et al. (2022); Tong et al. (2024) |
| Lane change | Highway driving | Hang et al. (2020); Larsson et al. (2021); Wang et al. (2023b); Xu et al. (2023); Zhao et al. (2024) |
| | Urban driving | Chang et al. (2023); Wang et al. (2021b) |
| | Two-lane road with large curvature | Yan et al. (2022) |
| | Not specific | Chen et al. (2024) |
| Merge | On-ramp merging | Hang et al. (2021, 2022c); Liu et al. (2024c); Nan et al. (2024); Schwarting et al. (2019); Toghi et al. (2021a, 2021b, 2022); Valiente et al. (2024); Wang et al. (2023b); Xue et al. (2023) |
| | Intersection merging | Wang et al. (2024b) |
| Overtaking | Urban driving | Lu et al. (2022); Zong et al. (2023) |
| | Highway driving | Hang et al. (2020, 2021); Zhao et al. (2024) |
| | Not specific | Wang et al. (2024b) |
| Highway exit | | Landolfi and Dragan (2018); Toghi et al. (2022); Valiente et al. (2024); Wang et al. (2023b) |
| Interact with pedestrian/Pedestrian collision avoidance | | Bhatt et al. (2022); Crosato et al. (2021); Crosato et al. (2023a); Pérez-Dattari et al. (2022); Sun et al. (2019); Taghavifar and Mohammadzadeh (2025) |
| Road cruising | | Wang et al. (2024b) |
| Platoon | | Wang et al. (2024a) |
| Bottleneck | | Peng et al. (2021); Xue et al. (2023) |
| Tollgate | | Peng et al. (2021) |
| Parking lot | | Peng et al. (2021) |
| Nudging parked cars on urban streets | | Bhatt et al. (2022) |
| Social occlusion inference | | Huang and Sun (2023) |
| Oncoming traffic | | Kolekar et al. (2020); Liu et al. (2024c) |
| Reacts to stalled car | | Landolfi and Dragan (2018) |
| Reacts to speeding | | Landolfi and Dragan (2018) |
| Reacts to ambulance | | Landolfi and Dragan (2018) |
| Car-following | | Kolekar et al. (2020); Larsson et al. (2021) |
| Social robot navigating | | Da and Wei (2023); Ferrer and Sanfeliu (2014); Galati et al. (2022); Hirose et al. (2024); Liu et al. (2020); Reddy et al. (2021) |
| Human trajectory forecasting | | Alahi et al. (2016); Gupta et al. (2018); Kothari et al. (2021); Kothari and Alahi (2023); Sadeghian et al. (2019); Vemula et al. (2018); Wang et al. (2024c) |
| Social reactions, feedbacks, and trust in AVs | | Joo and Kim (2023); Oliveira et al. (2019); Othman (2021); Schneble and Shaw (2021) |

Note: [a] Unsignalized intersection: Here, "unsignalized intersection" may include "unprotected left turn" or other scenarios (e.g., right-turning and going straight at unsignalized intersections), whereas the row of "unprotected left turn" is specifically about unprotected left turning through unsignalized intersections.

- What key efforts are necessary to drive the development and acceptance of automated vehicles?

The key insights derived from these expert interviews are summarized as follows.

With respect to the current practices and limitations of SCAVs, several critical shortcomings in the current generation of AVs have been identified:

- **Excessive conservatism**: Most current AVs often adopt overly defensive driving strategies, which may significantly compromise traffic efficiency.
- **Inability to interpret implicit communications**: Most current AVs struggle to decode subtle signals to understand the implicit "communications" from human drivers, such as waving hands or deceleration, which implies yielding right of way.
- **Challenges in adapting to various driving styles**: Most current AVs are unable to adapt effectively to various driving styles, especially the aggressive or assertive driving behaviors exhibited by surrounding HDVs.
- **Limited scenario anticipation**: Unlike human drivers, current AVs lack robust capabilities to foresee, anticipate, and prepare for dynamic future scenarios.
- **Cultural and normative inflexibility**: Current AVs are not yet designed to adapt their driving behaviors and styles to account for varying norms and driving cultures across different countries.

With respect to the research gaps and expectations, together with the literature review findings, the highlighted critical gaps and outlined priorities for advancing SCAVs are as follows:

- **Integration of sensing, planning and control**: Few studies connect AVs' sensing capabilities, particularly considering sensor inaccuracies, to trajectory planning and control. Given the importance of this in real-world deployment, more in-depth exploration is warranted.
- **Cultural and normative adaptation**: As limited research and development have incorporated cultural differences, driving norms, and implicit cues into automated driving models, this area deserves more attention.
- **Development of av communication pipelines**: There is a pressing need for AVs to express their intentions to other road users through, e.g., external human–machine interfaces (eHMIs) such as color-changing surfaces, signal lights, or LED panels on AVs.
- **AV-human mutual behavioral adaptation**: The long-term and short-term adaptations of human driver behavior when interacting with AVs and the corresponding adjustments AVs should make in response to those adaptations are seldom accounted for in the current development of AV driving models.
- **Network-wide and societal benefits**: Few studies have considered broader implications for overall network efficiency and societal benefits (e.g., total emissions across road networks) when different AV driving strategies, styles, and behaviors are deployed.
- **Interdisciplinary efforts**: Most research combines approaches from computer science, physics, mathematics, and engineering. Emerging efforts involving social psychology focus on adding concepts such as SVO, coordination tendencies, and courtesy. More advanced frameworks incorporating social psychology and other interdisciplinary

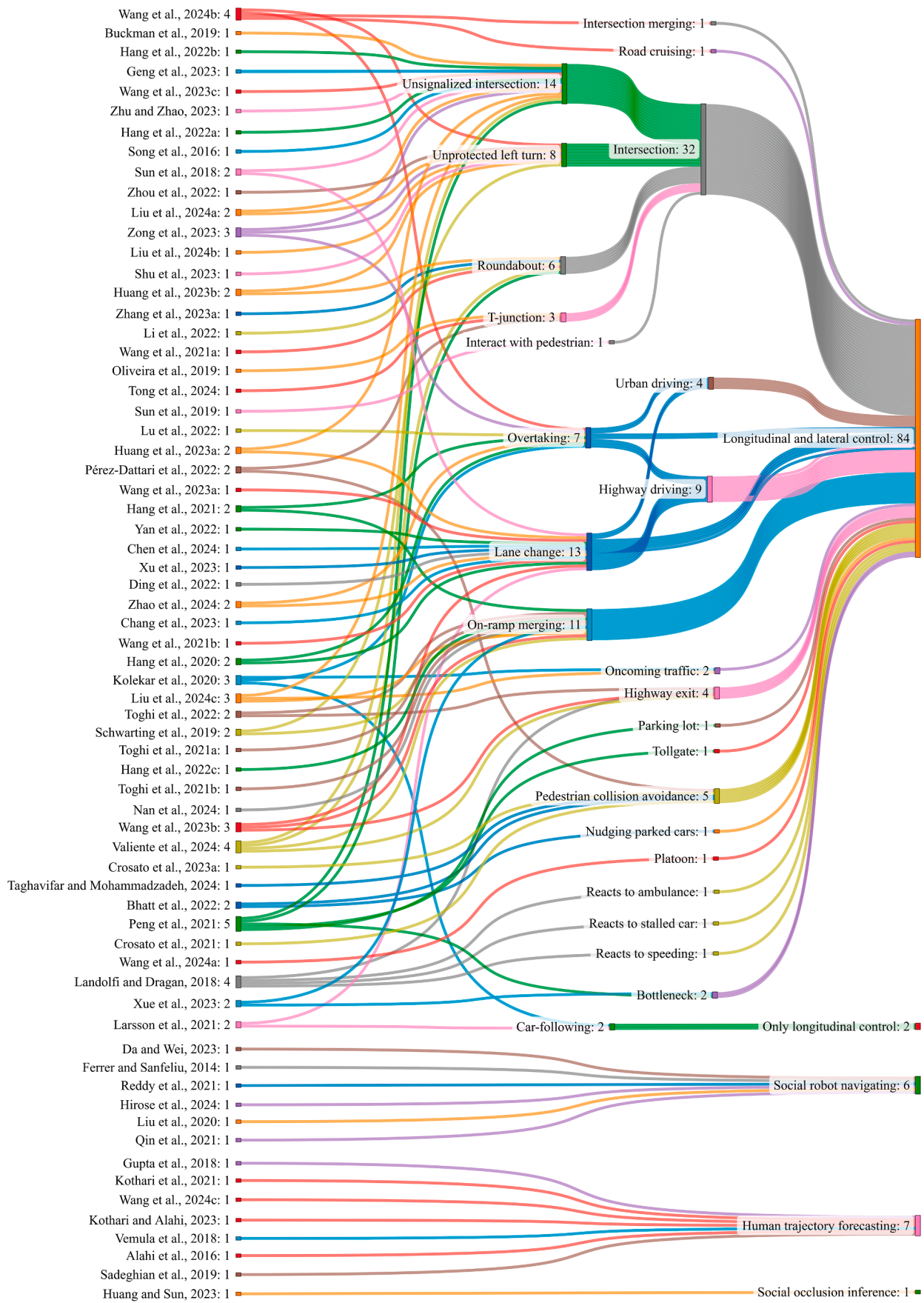

**Fig. 4.** Identified maneuvers involved in each study.

Note: A single paper may involve multiple maneuvers; thus, the total number of maneuvers can exceed the total number of reviewed papers (68).

fields are needed to deepen the understanding of human-AV interactions.

These insights constitute the basis for the conceptual framework in Section 3.2 to guide future research and development efforts in this area.

### 3.2. Proposed conceptual framework

By incorporating insights from the scoping review and addressing the identified gaps and research expectations from both the literature review and the expert interviews, a conceptual framework, as illustrated in Fig. 5, is proposed to guide future research and development on SCAVs.

Overall, this framework follows and adheres to the standard modular design for developing AVs, which includes sensing and perception modules, decision-making modules, planning modules, and control action modules. The differences and added values of the proposed conceptual framework are as follows.

1) Socially compliant decision-making module: The traditional decision-making module is enhanced, re-engineered, and transformed into the proposed socially compliant decision-making module. This modification integrates social components (including culture, norms, and cues), which may influence implicit interactions, and considers various driving styles (e.g., aggressive, cautious, and prosocial). The integration and embedding of these elements will help address the aforementioned limitations of cultural and normative inflexibility and challenges in adapting to various driving styles. Furthermore, the module incorporates mechanisms for bidirectional behavioral adaptation, enabling AVs to respond to human drivers' behavioral cues and adjust their responses accordingly, which will be illustrated later.

2) Safety constraint module: This module continuously monitors and enforces safety constraints to ensure that AVs operate within predefined safety boundaries. Although the socially compliant decision-making module should incorporate safety metrics, the dedicated safety constraint module serves as a critical safeguard, ensuring that all actions taken by the AV are within the safety limits, thereby preventing undesirable outcomes. The planning module in this framework encompasses both high-level path planning and behavior planning (e.g., lane changes, merging) as well as low-level motion planning (e.g., longitudinal and angular velocity, acceleration), all of which must adhere to the safety constraints outlined by this module.

3) Trade-off between ego and network-level benefits: A fundamental challenge (which is currently missing) in AV development is balancing the individual benefits of the ego vehicle (such as safety, comfort, and efficiency) with the broader benefits to the road network and other road users. The proposed framework emphasizes the necessity of managing this trade-off (as shown in the utility components within Fig. 5), acknowledging that optimal performance for individual vehicles should not come at the expense of overall network efficiency or societal benefits. This trade-off should be managed dynamically, on a case-by-case basis, to ensure a balanced approach that maximizes both individual and collective outcomes (i. e., a more holistic, systems-level perspective). This requires close collaboration between AV developers, road operators, and regulatory authorities to align objectives and responsibilities. By managing the trade-off adaptively, this module helps meet the aforementioned expectations regarding network-wide and societal benefits.

4) Bidirectional behavioral adaptation module: A key novel contribution of the proposed framework is the introduction of a bidirectional behavioral adaptation module. This module addresses the phenomenon where human drivers adapt their behavior in response to the presence and actions of AVs in mixed traffic. For example, drivers may exploit the defensive behavior of AVs by engaging in more aggressive driving when interacting with them. To mitigate this, AVs must adapt their behaviors in return, effectively responding to changes in human driving patterns and fostering more balanced and cooperative interactions. The module is designed to facilitate a dynamic, iterative process of mutual adaptation, wherein both AVs and human drivers adjust their actions to optimize safety, traffic flow, and overall road network efficiency under mixed-traffic conditions. For successful real-world deployment, the bidirectional behavioral adaptation module must undergo continuous updates, both in the short-term and long-term, to account for evolving traffic conditions and varied human driving behaviors. This ensures that the module remains responsive to a wide array of scenarios, thereby supporting the integration of AVs into diverse traffic contexts. This module helps alleviate the aforementioned limitations of excessive conservatism and challenges in adapting to various driving styles and helps meet the expectations of AV-human mutual behavioral adaptation.

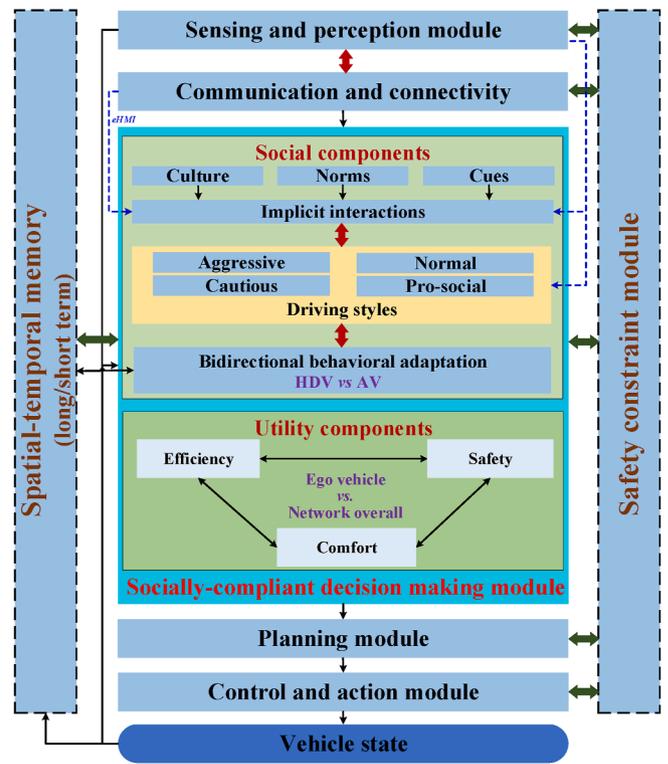

**Fig. 5.** Proposed conceptual framework for developing socially compliant AVs.

5) Spatial-temporal memory module: The spatial-temporal memory module is designed to facilitate the long- and short-term updating of knowledge and driving rules, as well as to increase awareness of ongoing behavioral adaptations. This module enables AVs to incorporate historical interaction data and adapt their decision-making strategies over time. By maintaining a dynamic memory of past interactions, AVs can continuously refine their understanding of human-AV dynamics, ensuring that driving strategies incorporate lessons learned from prior experiences. This module is essential for the effective integration and implementation of bidirectional behavioral adaptation within the broader AV decision-making framework.

Explanations regarding the remaining limitations, gaps, and expectations presented in Section 3.1 are as follows.

The limited scenario anticipation is addressed by the sensing and perception module as well as the communication and connectivity module, which forms the backbone of the framework's ability to predict and respond to dynamic traffic scenarios. As illustrated by the dashed arrows in Fig. 5, the socially compliant decision-making module depends on seamless integration with advanced sensing and communication systems. Sensing technologies, including cameras, LiDAR, and radar, deliver real-time data on the positions, velocities, and behavioral cues (e.g., accelerating, decelerating, and braking patterns) of surrounding road users. These data enable the socially compliant decision-making module to interpret social norms, anticipate interactions, and estimate and adapt to diverse driving styles. In addition, communication and connectivity systems such as vehicle-to-vehicle (V2V), vehicle-to-infrastructure (V2I), vehicle-to-everything (V2X), and eHMI provide critical supplementary inputs, such as the signaled intentions or planned trajectories of other vehicles. Together, these systems enhance the AV's situational awareness, ensuring that social decisions are informed by a comprehensive understanding of the traffic environment, thereby mitigating limited scenario anticipation and grounding the framework in operational reality.

The inability to interpret implicit communications can be alleviated through the proposed eHMI, which connects the sensing and perception module to the element of implicit interactions within the socially compliant decision-making module (shown in blue text and dashed arrows in Fig. 5). The eHMI allows AVs to convey their intentions (such as yielding or lane-changing) more effectively to other road users, facilitating mutual understanding and smoother interactions in mixed-traffic settings. This will also help meet expectations regarding the development of AV communication pipelines, fostering improved communication between AVs and surrounding HDVs, pedestrians, and cyclists.

Additionally, the integration of sensing, planning, and control relies on an advanced, robust sensing and perception module capable of managing uncertainties and sensing failures. While such a module is integral to the framework's success, developing cutting-edge sensing and perception techniques exceeds the scope of this study and remains a broader research challenge itself, warranting further exploration.

Finally, while vehicle connectivity, including V2V, V2I, and V2X communication with pedestrians, cyclists, and other road vehicles, plays a vital role in addressing several limitations and gaps, it is important to clarify that these aspects are beyond the scope of this study. Connectivity is recognized as a crucial element in the broader ecosystem of autonomous driving, meriting its own dedicated line of research. Owing to space limitations, this study could not delve deeply into that area.

## 4. Online questionnaire survey

To evaluate and verify the proposed framework for developing SCAVs, an online questionnaire-based survey was conducted. The survey was disseminated via targeted email distribution lists, including those of relevant expert groups such as the Universities' Transport Study Group (UTSG) and the TRAIL Research School. Additionally, the survey was actively promoted during key academic conferences, including the IEEE Intelligent Transportation Systems Conference (ITSC) and the IEEE Intelligent Vehicles Symposium (IV). The participants were asked to answer a sequence of questions, including multiple-choice questions, rank-order scale questions, rating scale questions, and open-ended questions. The questions are presented in seven subsections. The online survey took approximately 15 min to complete. The survey can be accessed at https://lnkd.in/evg6Dn9W. To promote expert and professional participation in the survey, it was mentioned that every successful and qualified response would result in a five-euro donation to the United Nations Road Safety Fund (https://roadsafetyfund.un.org/). The survey questionnaire is provided in full for reference in Supplementary Attachment 2 at https://lnkd.in/gpceU6gQ.

### 4.1. Respondent profile

A total of 99 responses were collected from experts across various nations and continents. Nine responses were excluded from the analysis because of contradictions in the answers or because the respondents self-identified as lacking confidence in their responses. Thus, 90 responses from experts were included in the final analysis. These experts represent a diverse range of roles in professional services, including researchers from universities, research institutes, and industry companies; developers from OEMs; policymakers; consultants; technicians; and professional drivers.

Fig. 6 illustrates the distribution of the respondents' profiles. The remaining 90 respondents were from 29 countries across 6 continents. The majority of the respondents were from European countries and China, reflecting substantial representation in this study. Notably, only one respondent originates from the USA, a leading hub for AV technology development and deployment, resulting in its inclusion within the category of "Other" in Fig. 6a. Despite this limited presence of the USA, China's significant participation aligns with its own prominence in AV innovation and deployment, enriching the study with valuable insights from a key market.

All 90 respondents claimed to be familiar with the concept and technology of automated vehicles, and more than half (54 out of the 90) of them were working in a field directly related to automated vehicles. Among them, 35 respondents are involved in developing AVs, 8 are engaged in testing automated driving functions, 3 of them are qualified safety/test drivers, and 1 is researching human factors related to AVs.

In terms of professional roles, 49 respondents are researchers, 18 are consultants, 7 are policymakers, and 2 are developers or technicians at OEMs. Notably, one respondent claimed to be an associate editor for a relevant journal, one claimed to be responsible for the implementation of vehicle regulations by public authorities, and another worked on the national strategy for the deployment of AVs. Furthermore, 86 out of the 90 respondents held a driving license, with 6 claiming to have a professional driving qualification. These findings underscore the diverse expertise and perspectives that the respondents bring to the survey, enhancing the credibility of the survey results.

### 4.2. Benefits of SCAVs and willingness to purchase or use them

With respect to the benefits of SCAVs, participants were asked to rate the extent to which they thought that SCAVs would influence overall traffic safety and efficiency. The rating is based on a 7-point Likert scale with "−3" meaning strongly worsening, "0" standing for neutral/no influence, and "3" indicating strongly improving. As demonstrated in Fig. 7, the majority of respondents believe that SCAVs contribute positively to both overall traffic safety and efficiency.

The average rating for potential improvement in safety is 1.04, whereas the average rating for efficiency is 0.54. These figures indicate that, on average, respondents perceive SCAVs as having greater potential to enhance safety than to improve efficiency, but both are seen as contributing positively.

Similarly, participants were asked about their willingness to purchase SCAVs when considering a vehicle purchase or their willingness to use them for on-demand mobility services while traveling. The majority responded positively, as shown in Figs. 8 and 9. Specifically, 72 respondents indicated that they would like to buy an SCAV, whereas only 8 stated that they would never consider purchasing one, even if such AVs were less expensive.

Furthermore, acknowledging the suitability of SCAVs for shared on-demand travel services, 77 respondents expressed a willingness to use them for trips, whereas only 4 indicated that they would not use SCAVs, even if they were more affordable.

Notably, some participants emphasized that they prioritized functionality and performance over price, expressing a preference for public transport or other sustainable options that meet their specific needs; thus, they were categorized into the "Other" group.

### 4.3. Development of SCAVs

#### 4.3.1. Rating and ranking of the identified key technical capabilities

In the context of developing SCAVs, experts' opinions on the importance of various technical aspects required for AVs to exhibit socially compliant behaviors were assessed. Corresponding to the developed conceptual framework (Fig. 5), the respondents were asked to rate 9 key technical capabilities on a scale from 1 to 7, where 1 represented "Least Needed" and 7 represented "Strongly Needed". The evaluated technical aspects were as follows:

- **Anticipation capability:** The ability to anticipate the intended actions of other road users;
- **eHMI communication capability:** The ability to convey intended actions effectively through external human–machine interaction (eHMI);
- **Social and cultural alignment:** The ability to adapt to different local cultures, social norms, and cues;
- **User acceptance:** The ability to consider acceptance levels among drivers, passengers, and nearby road users;
- **Driving style adaptation:** The ability to adjust to varying driving styles of surrounding human drivers, such as aggressive or defensive and prosocial or egoistic;

- **Bidirectional behavioral adaptation:** The ability to enable mutual adaptation between AVs and human drivers over time;
- **Multi-objective optimization:** The ability to balance multiple goals, such as safety, efficiency, energy consumption, and environmental impact;
- **Trade-off management:** The ability to maintain trade-offs between the AV's benefits and those of surrounding traffic participants and between the ego AV's benefits and benefits at the network (regional) level;
- **Spatial-temporal memory buffer integration:** Spatial-temporal memory buffers (short-, medium-, and long-term) are incorporated to refine driving strategies continually.

As shown in Fig. 10, the respondents rated the extent to which they believed these properties should be integrated into SCAVs. All 9 key technical capabilities were rated as significant, with average ratings exceeding 4.8, which supports and verifies the elements proposed in the conceptual framework (Fig. 5). Their ratings did not vary much, with anticipation capability receiving the highest average rating (6.29), followed by the capabilities of multi-objective optimization (5.76) and trade-off management (5.61).

As demonstrated in Fig. 11, the respondents were also asked to rank the top 3 most important aspects among the 6 selected capabilities in terms of medium-term development (1–3 years), supposing that there are limited resources for developing SCAVs. The ranking results indicated that anticipation capability ranked first, followed by multi-objective optimization, which is consistent with the results shown in Fig. 10.

Furthermore, as illustrated in Fig. 12, respondents were asked to rank the top 2 most important aspects among the 4 selected capabilities for long-term development (in the next 5–10 years or longer), again assuming limited resources for developing SCAVs. The results revealed that bidirectional behavioral adaptation ranked first, followed by spatial-temporal memory buffer integration, which is reasonable and aligns well with the proposed conceptual framework in Fig. 5.

These ratings and rankings yield critical insights into which technical features are deemed essential and urgent for enabling AVs to navigate complex social interactions effectively. Such data-driven insights will be invaluable in guiding the prioritization and future technical development of SCAVs.

### 4.3.2. Rating the possibility of mathematically modeling the identified key technical capabilities

With respect to the implementation of the identified key technical capabilities, the respondents were asked to rate the possibility and feasibility of mathematically modeling the six identified key technical capabilities of social and cultural alignment, driving style adaptation, bidirectional behavioral adaptation, multi-objective optimization, trade-off management, and spatiotemporal memory buffer integration. Ratings were provided on a scale from 1 to 7, where 1 represented "Not Possible" and 7 represented "Highly Possible". The results are depicted in Fig. 13.

All the examined 6 key technical capabilities were found to be feasible for mathematical modeling. Multi-objective optimization was rated and deemed the most feasible, followed by trade-off management, which is expected given that both of them could be modeled as typical optimization problems. In contrast, social and cultural alignment was identified as the most challenging and least feasible for mathematical modeling, followed by bidirectional behavioral adaptation and spatial-temporal memory buffer integration, which is also reasonable. This aligns with earlier recommendations for interdisciplinary cooperation, particularly drawing on knowledge, and insights from social psychological domains alongside advancements in computer science.

### 4.3.3. Suggestions from the respondents

The respondents were invited to share suggestions and insights through open-ended questions such as "What else would you expect for

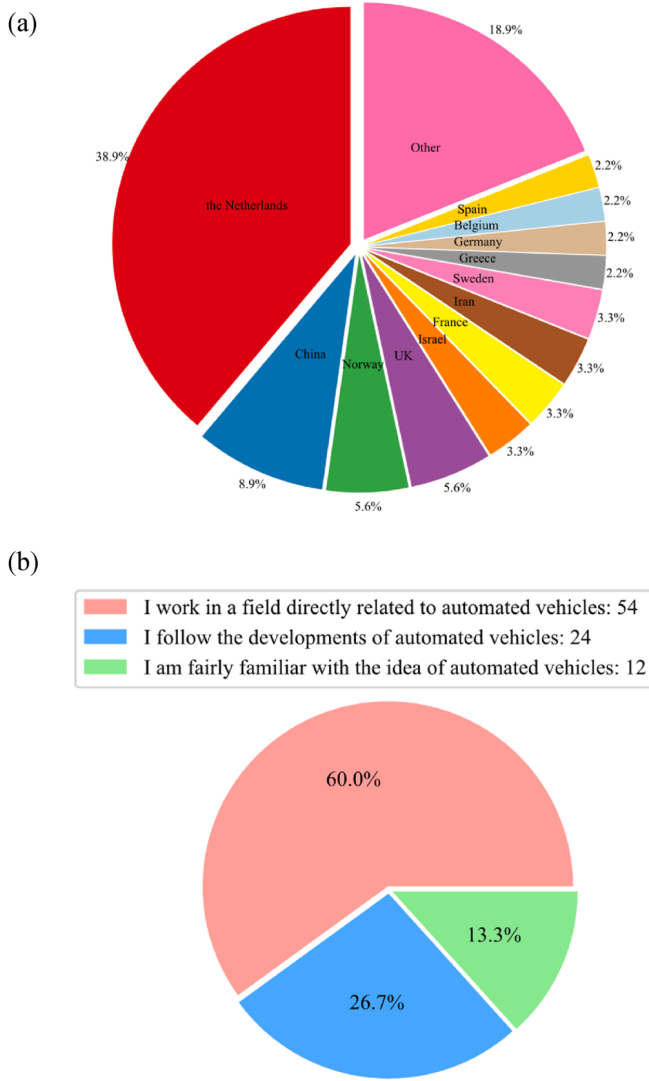

**Fig. 6.** Distribution of respondents' profiles: (a) Residence countries and (b) familiarity with AVs.
Note: The "Other" category in (a) encompasses respondents from 17 countries (out of 29 total) not individually listed, including USA, Canada, Australia, Italy, and India, among others, in addition to the 12 explicitly named nations (the Netherlands, China, Norway, UK, Israel, France, Iran, Sweden, Greece, Germany, Belgium, Spain). Notably, USA, a key AV technology hub, is grouped under "Other" due to its minimal representation (only one respondent post-preprocessing), which is insufficient for a distinct category.

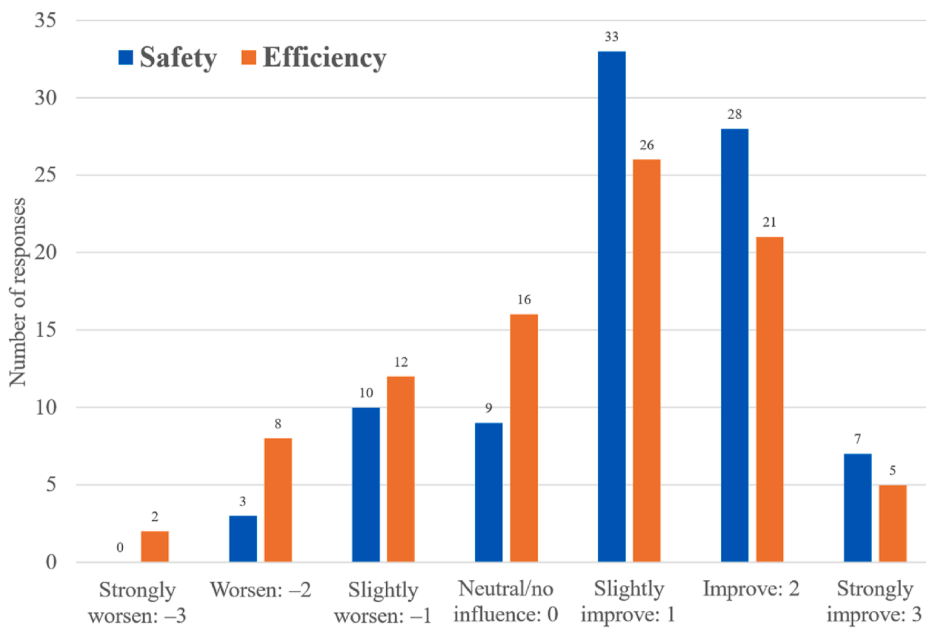

**Fig. 7.** Rating distribution of the extent to which the participants thought that SCAVs would influence overall traffic safety (light blue) and efficiency (orange).

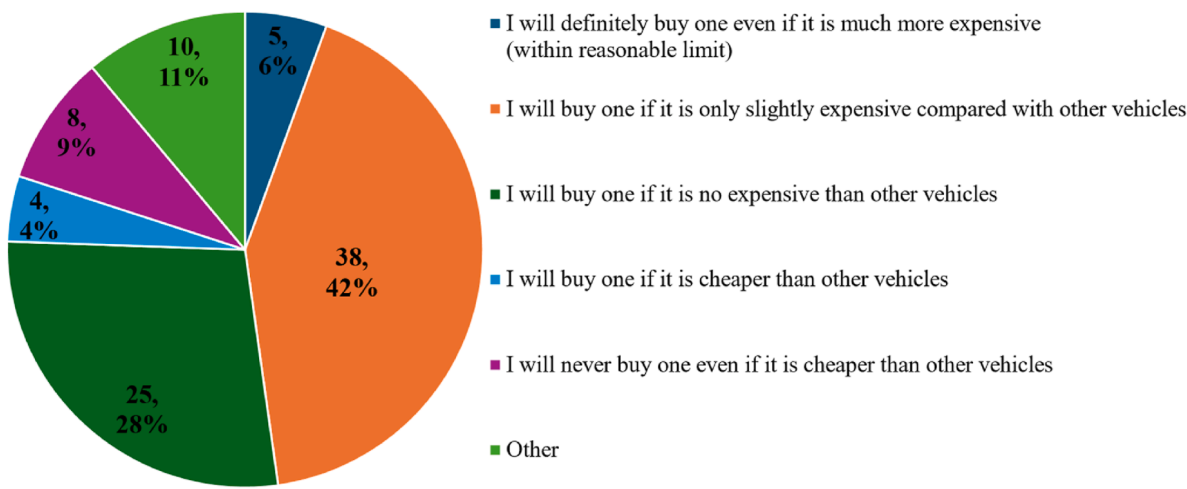

**Fig. 8.** Distribution of willingness to buy one socially compliant automated vehicle.
Note: The "Other" category includes special responses beyond predefined options, e.g., preferences for cycling, walking, or public transit explicitly rejecting car ownership.

the socially compliant automated vehicles?" and "Do you have any further comments for better development of socially compliant automated vehicles?" A range of thoughtful responses were collected, which, after in-depth analysis, were summarized, further upgraded, and polished as follows:

The development of SCAVs must prioritize safety and trust as core principles. Safety should remain paramount across all stages of development, and building trust between humans and SCAVs requires transparency, effective trust modeling, and clear communication of the vehicle's decision-making processes and intentions to its users and other road participants. The respondents emphasized the need for ML models to be trained via curated, unbiased datasets that reflect socially responsible driving behaviors rather than exceptional cases such as those of professional drivers (e.g., F1 pilots). Additionally, initial deployment should focus on less complex environments, such as highways and provincial roads, before progressing to urban settings, where social compliance becomes more intricate and essential.

Infrastructure upgrades are also vital to support the successful deployment of SCAVs. This includes the development of dedicated AV lanes, V2I and V2X communication networks, and robust systems with reliable backup mechanisms to prevent failures in smart traffic management systems. The respondents also highlighted the importance of balanced policy frameworks that encourage shared mobility solutions, such as controlled fleets of robotaxis, over private ownership of AVs. Collaboration among OEMs, regulators, and other stakeholders is deemed critical for fostering open communication, pooling knowledge, and advancing technical priorities strategically.

An interdisciplinary and culturally sensitive approach is needed to reflect the diversity of societal needs in SCAV behaviors. Human factors must be central to design, ensuring that AVs can adapt to the social norms and behaviors of both drivers and other road users, such as cyclists and pedestrians, who are often overlooked. SCAVs should strike a balance between idealized performance and relatable, realistic behaviors that align with the imperfect nature of human driving.

Ultimately, the success of SCAVs hinges on the careful prioritization of technical and social efforts, given the significant time and resources required for development. Transparent AI systems, robust infrastructure, and a focus on public acceptance and trustworthiness will be pivotal in ensuring SCAVs' seamless integration into society. These vehicles must not only navigate the immediate social and cultural contexts of their operation but also anticipate the long-term challenges of mixed-traffic environments and future scenarios dominated by automation. With thoughtful design and strategic planning, SCAVs can deliver safe, reliable, and socially aligned mobility solutions that meet the evolving needs of diverse communities.

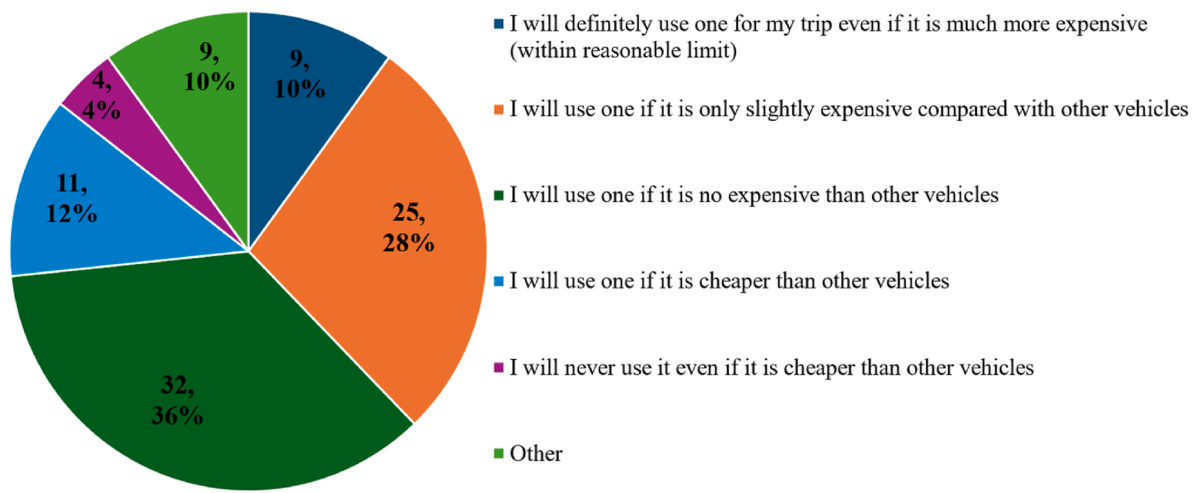

**Fig. 9.** Distribution of willingness to use socially compliant automated vehicles for trips.
Note: The "Other" category covers special responses outside listed options, e.g., bus travel, cycling, or car-sharing, instead of car use.

Furthermore, as the deployment of AVs becomes increasingly widespread, a growing body of empirical evidence on real-world AV behavior is emerging. This provides a valuable opportunity to investigate not only how AVs interact with human-driven vehicles but also how they respond to each other. Understanding interactions both within the same brand and between different brands of AVs is an area that remains underexplored but is critical for fostering interoperability, social compliance, and collaborative traffic systems. Such studies could reveal how variations in algorithms, decision-making priorities, and communication protocols influence the dynamics of AV interactions. By fostering cross-brand standardization and promoting cooperative driving behaviors among AVs, the industry can take a significant step toward realizing the vision of a harmonized, intelligent transportation system that benefits all road users. Expanding research in this direction would further support the development of SCAVs that are not only socially compliant but also capable of thriving in increasingly complex and automated traffic environments.

## 5. Conclusions, limitations, and future research

This study represents the first comprehensive scoping review of the current state of the art in the development of SCAVs, systematically identifying key concepts, methodological approaches, and research gaps in the field. Through a rigorous review of the literature and expert interviews, this study elucidated critical pain points and research gaps while outlining vital research expectations essential for advancing SCAV development. Building on these insights, this study proposes a novel conceptual framework designed to address the multifaceted and interdisciplinary challenges of SCAVs in mixed-traffic environments. The framework outlines the key capability elements necessary for SCAVs and incorporates crucial considerations across technical, social, and cultural dimensions, effectively bridging theoretical insights with practical applications to achieve socially compliant automation.

To validate the conceptual framework, an online questionnaire-based survey was conducted, confirming the relevance of the framework's key elements and technical capabilities. Among these, anticipation capability has emerged as the most significant and urgent requirement for midterm implementation (1–3 years), reflecting its importance in enabling SCAVs to predict and adapt to dynamic road scenarios, especially with respect to interactions with HDVs. For long-term development (5–10 years or more), bidirectional behavioral adaptation—the ability to dynamically and mutually interact with and learn from other road users—and spatial–temporal memory buffer integration were identified as the most critical priorities. These findings offer actionable insights for research and development (R&D) in both academia and industry, serving as a strategic roadmap for integrating social compliance into automated driving systems. They highlight research priorities and guide the creation of SCAVs that align with societal expectations. For researchers, the proposed conceptual framework identifies focus areas and key elements to be studied. For the industry, it provides actionable insights into developing and embedding social compliance in AV systems, enabling scalable and context-sensitive deployment. The developed framework can also foster collaboration among academia, industry, and policymakers; ensure that technical innovation aligns with societal needs and regulatory standards; accelerate the path toward SCAV; and further promote safe and socially inclusive automated mobility solutions.

By providing a structured and interdisciplinary approach, this study contributes to the foundation of socially aware and ethically aligned AV technologies, laying the groundwork for safe, reliable, and socially compliant automated mobility solutions.

Despite its meaningful contributions, this study has several limitations that provide opportunities for further research. First, in the scoping review, as mentioned above, the scoping review did not analyze or summarize in detail the experiments, model performance, or results from the reviewed studies. Furthermore, the study did not thoroughly investigate scenarios involving multivehicle interactions, particularly among multiple AVs. As AV penetration rates increase, understanding these interactions becomes critical. Future reviews could address these gaps to provide a more comprehensive assessment of current research in this field.

Second, while the study emphasized the importance of anticipation capability, it did not extensively address its relationship with perception, particularly perception under uncertainty. This critical aspect, which includes managing ambiguous or incomplete information in real-world scenarios, represents a highly complex research domain that warrants dedicated research attention. The development of robust perception systems that can handle uncertainties will significantly enhance SCAVs' ability to navigate and interact socially in diverse and

(a)

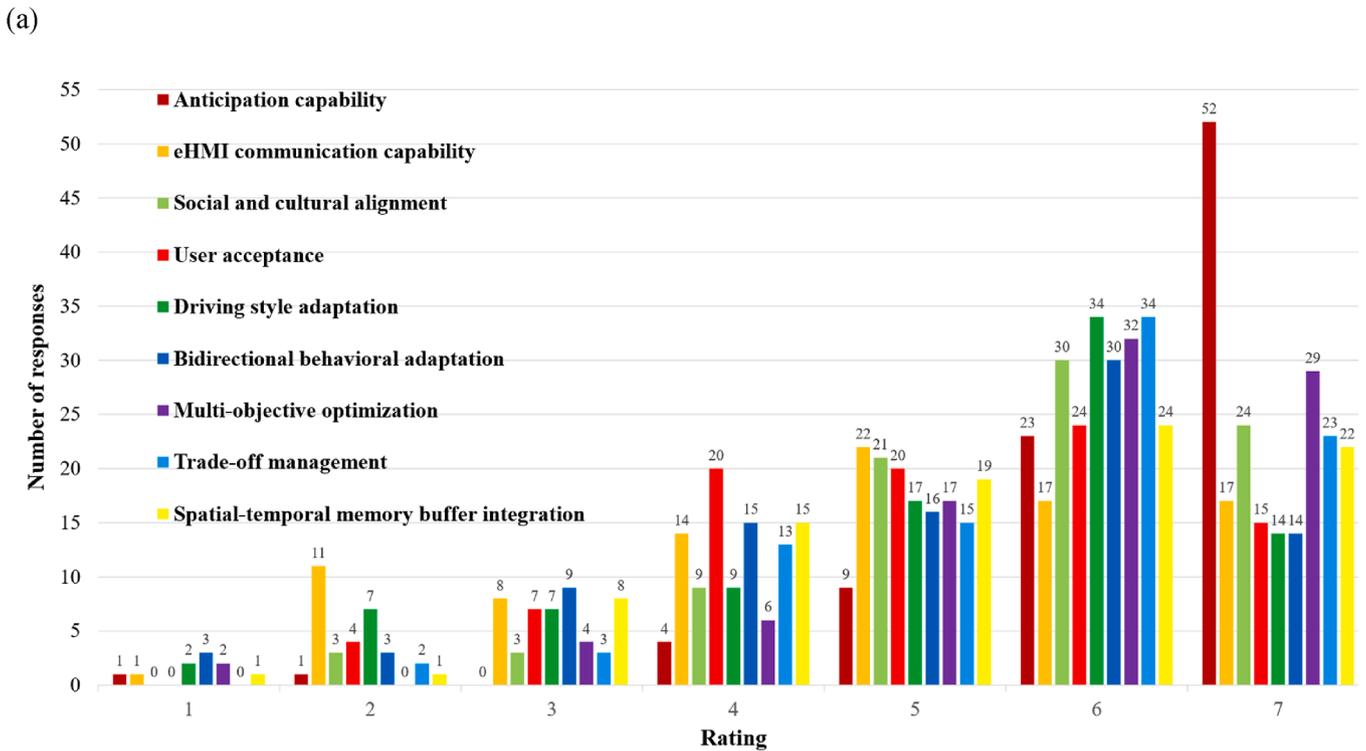

(b)

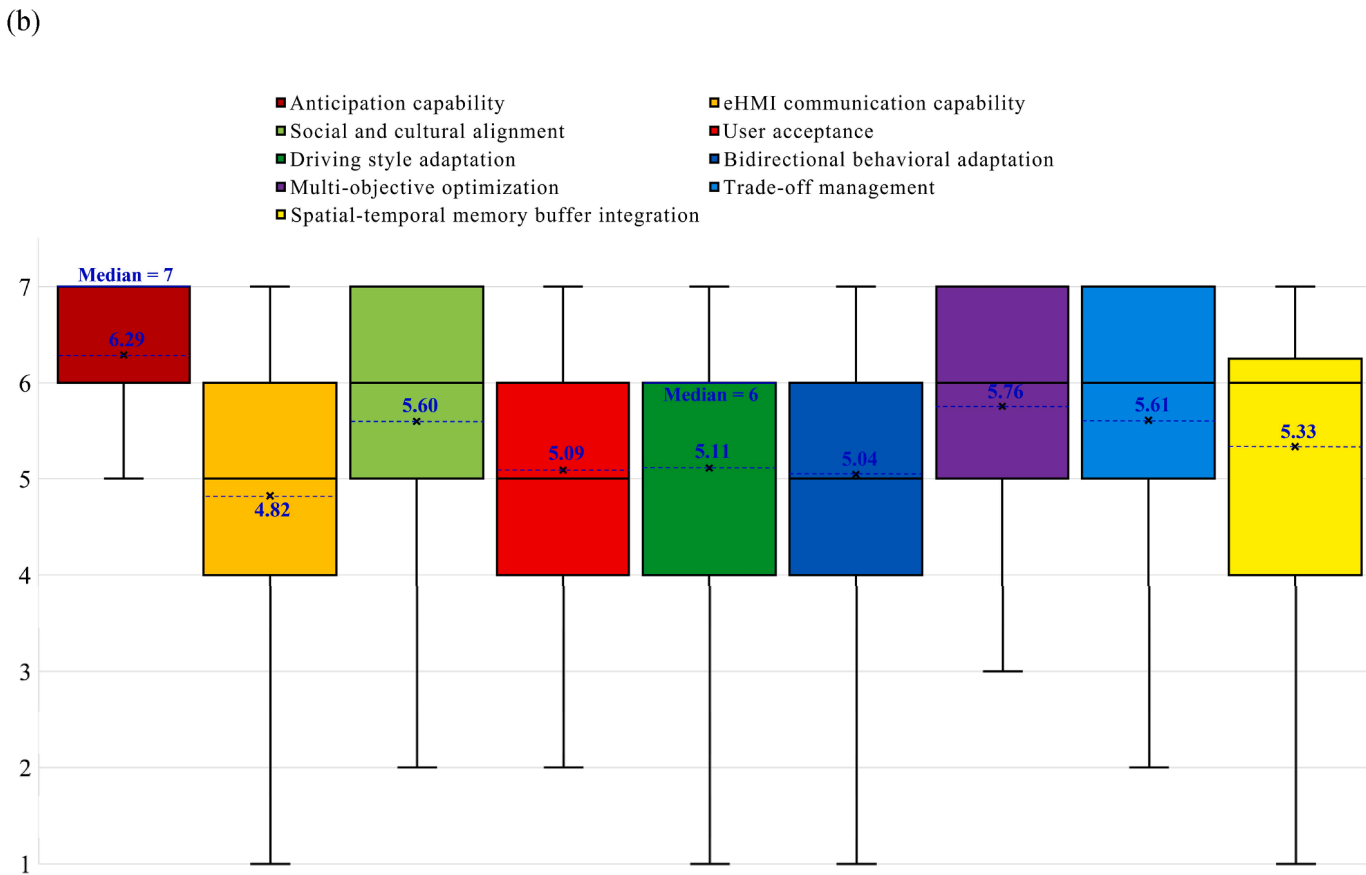

**Fig. 10.** Ratings of the importance of 9 key technical capabilities for developing SCAVs: (a) detailed rating distributions for each capability and (b) boxplot of the rating scales for each capability.

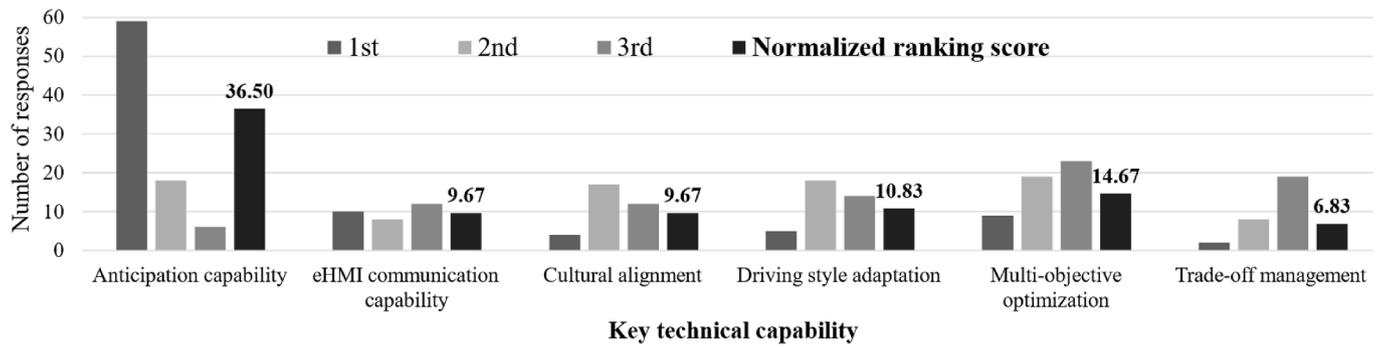

**Fig. 11.** Ranking results for 6 selected technical capabilities regarding their priorities for developing socially compliant AVs in the medium term.

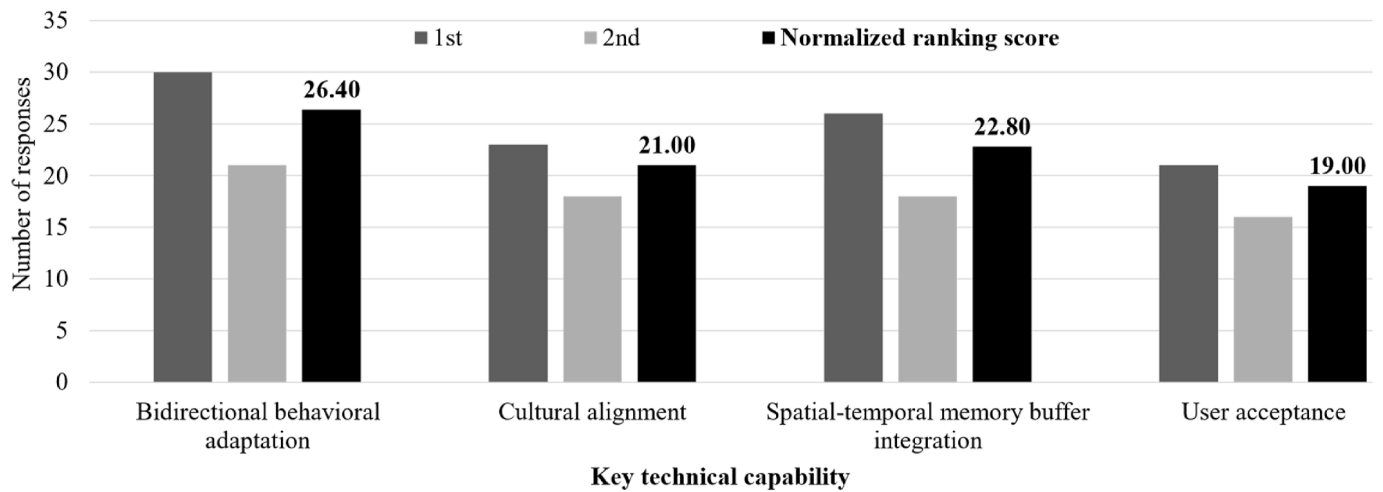

**Fig. 12.** Ranking results for 4 selected technical capabilities regarding their priorities for developing socially compliant AVs in the long term.

unpredictable environments. Similarly, connectivity, although recognized as an essential enabler, has not been explored in depth. Future work could delve into the integration and benefits of V2X technologies to support seamless communication between AVs, infrastructure, and road users for SCAV development.

Third, the study did not extensively examine interactions between AVs and vulnerable road users, such as cyclists and pedestrians. These interactions are crucial for ensuring that SCAVs can operate safely and effectively in complex urban environments, where unpredictable behavior from such road users often creates additional challenges. Addressing this limitation will not only enhance SCAVs' ability to anticipate and respond to the movements of vulnerable road users but also foster greater public trust in and acceptance of AV technologies. Such efforts will help make SCAVs more inclusive and adaptable to diverse road user types, ultimately contributing to safer and more equitable urban mobility systems.

Additionally, while our framework is designed to be adaptable to mixed-traffic environments broadly, it does not explicitly investigate how social behaviors vary across specific settings, such as urban, rural, and campus environments. Urban areas may require SCAVs to prioritize frequent, short-range interactions with diverse road users, whereas rural settings might involve adapting to less structured roads and unpredictable behaviors. With their mix of pedestrians, bicycles, and vehicles in confined spaces, campus environments could demand unique navigation strategies. Future research should explore these environmental differences to refine and validate our framework, tailor social compliance strategies to context-specific challenges and enhance the generalizability of SCAVs.

Moreover, the geographic distribution of respondents in the online survey of this study is predominantly concentrated in European countries and China, with only one participant from USA. This imbalanced representation may introduce potential cultural and contextual biases into the study's findings. Social compliance in driving behaviors is influenced by regional norms, regulations, and infrastructure designs, meaning that the current sample may not fully reflect global perspectives. Although China's substantial participation aligns with its prominence in AV development and deployment, the near absence of respondents from USA, another global leader in this domain, may underrepresent critical insights from a major AV market, thereby limiting the generalizability of the findings. This imbalance could particularly affect perceptions of social compliance expectations across diverse regional contexts. Future research should prioritize a more balanced sample, increasing representation from key AV markets such as USA to encompass diverse technological and cultural viewpoints.

Finally, this study does not address the broader systemic challenges of integrating SCAVs into existing infrastructure and ecosystems. Factors such as regulatory alignment, public acceptance, and economic feasibility remain critical to the successful deployment of SCAVs and must be explored further. In particular, balancing the needs of private and shared ownership models, addressing the environmental impact of SCAVs, and mitigating potential socioeconomic disparities should form part of future interdisciplinary research efforts.

For future research, a significant barrier to SCAV research is the scarcity of real-world field data, which restricts much of the current literature to simulation-based methodologies. Although simulations provide a controlled setting for modeling social compliance, they struggle to capture the full spectrum of human unpredictability. This shortfall limits the validation of SCAV frameworks in authentic mixed-traffic contexts, potentially leading to overestimated performance and overlooked vulnerabilities. Overcoming this requires prioritizing empirical field data collection, potentially through partnerships with AV testing initiatives (e.g., industry-led trials or regulatory pilot programs) or by harnessing data from controlled urban deployments. While such endeavors are resource intensive, they are crucial for transitioning SCAV solutions from theoretical constructs to reliable, real-world applications, thereby significantly increasing their practical robustness and relevance.

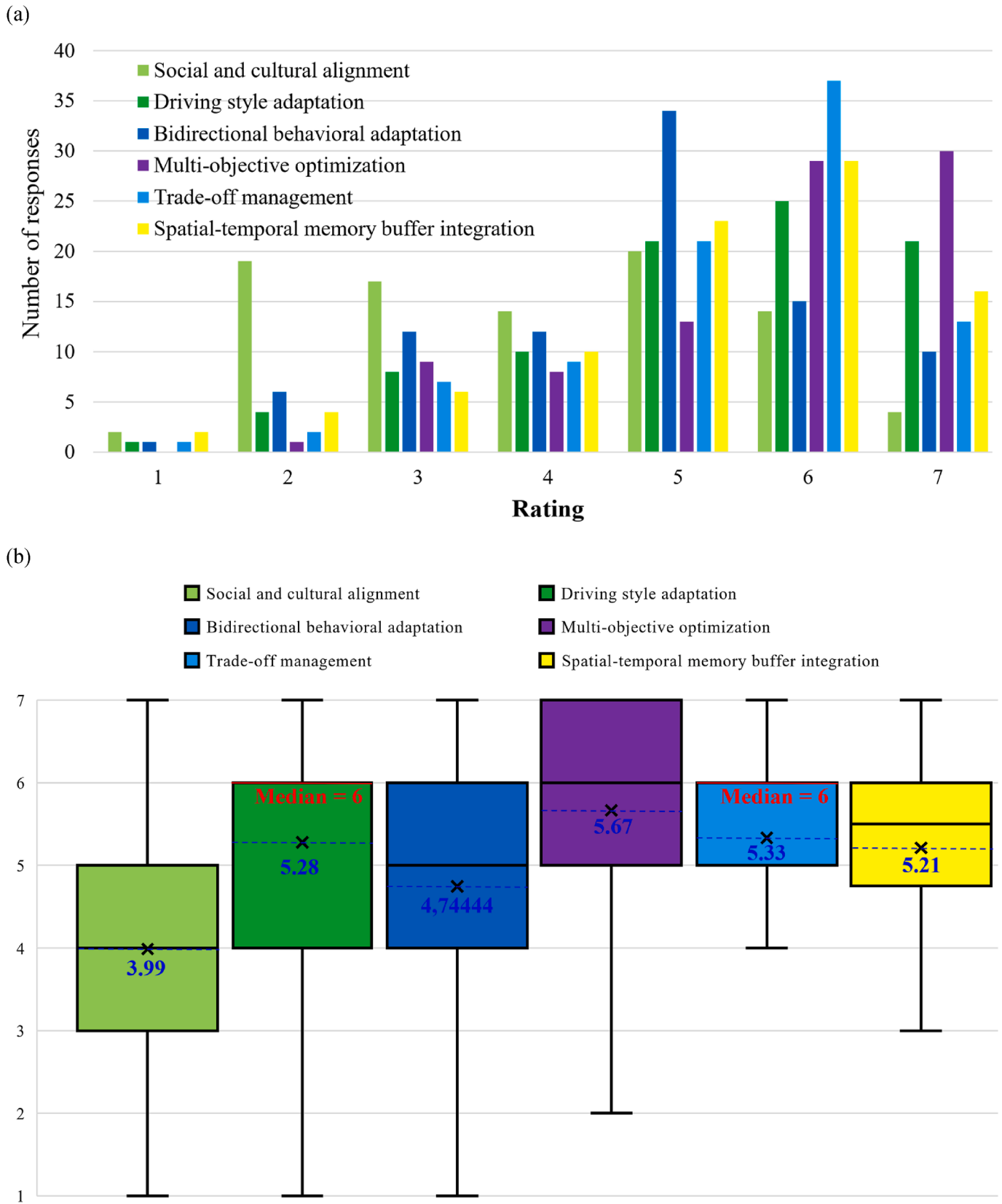

**Fig. 13.** Ratings on the feasibility of mathematically modeling the 6 identified key technical capabilities for developing socially compliant automated vehicles (AVs): (a) detailed rating distributions for each selected capability, (b) boxplot of the rating scales for each capability.

In conclusion, while this study provides a valuable foundation for SCAV development, it highlights the complexity and interdisciplinary nature of the challenges ahead. By addressing the identified limitations and advancing research in these critical areas, future efforts can build on the insights and framework presented here to create SCAV systems that are not only technically advanced but also socially responsible and globally inclusive.

## CRediT authorship contribution statement

**Yongqi Dong:** Visualization, Investigation, Conceptualization, Writing – original draft, Methodology, Data curation, Writing – review & editing, Validation, Formal analysis. **Bart van Arem:** Writing – review & editing, Conceptualization, Resources, Supervision. **Haneen Farah:** Supervision, Funding acquisition, Validation, Project administration, Writing – review & editing, Resources, Conceptualization.

## Replication and data sharing

The anonymized data and accompanying code used in this research are available at https://doi.org/10.4121/3a46e61c-f5f0-4399-a4b8-4d146b62a4f7.

## Declaration of competing interest

The authors declare that they have no known competing financial interests or personal relationships that could have appeared to influence the work reported in this paper.

## Acknowledgements

This work was supported by Applied and Technical Sciences (TTW), a subdomain of the Dutch Institute for Scientific Research (NWO) through the Project Safe and Efficient Operation of Automated and Human-Driven Vehicles in Mixed Traffic (SAMEN) under Contract 17187.

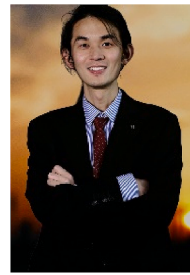

**Yongqi Dong** received the B.S. degree in telecommunication engineering from Beijing Jiaotong University, Beijing, China, in 2014; the M.S. degree in control science and engineering from Tsinghua University, Beijing, China, in 2017; and the Ph. D. degree in transport and planning from Delft University of Technology, Delft, the Netherlands, in 2025. He is currently working as a researcher and group leader with the Chair of Highway Engineering, RWTH Aachen University. His research interests include deep learning, transportation big data, automated driving, and traffic safety. He sought to employ artificial intelligence and interdisciplinary research as tools to shape a better world.

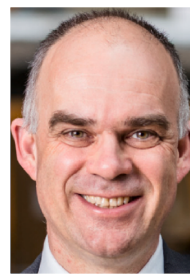

**Bart van Arem** received the M.S. and Ph.D. degrees in applied mathematics from the University of Twente, Enschede, the Netherlands, in 1986 and 1990, respectively. From 1992 to 2009, he was a researcher and the program manager with TNO, working on intelligent transport systems, in which he has been active in various national and international projects. Since 2009, he has been a Full Professor of intelligent transport modeling with the Department of Transport and Planning, Faculty of Civil Engineering and Geosciences, Delft University of Technology, Delft, the Netherlands. His research interests include modeling the impact of intelligent transport systems on mobility.

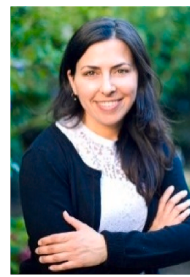

**Haneen Farah** is an Associate Professor at the Department of Transport and Planning and the Co-Director of the Traffic and Transportation Safety Lab, Delft University of Technology. She received the Ph.D. degree in transportation engineering from the Technion-Israel Institute of Technology. Before joining TU Delft, she was a postdoctoral researcher at the KTH Royal Institute of Technology, Stockholm, Sweden. Her research interests include road infrastructure design, road user behavior, and traffic safety. She is currently examining the implications of advances in vehicle technology and automation for road infrastructure design and road user behavior within the framework of several national and international projects.